\pdfoutput=1

\documentclass[letterpaper, 10 pt, journal, twoside]{IEEEtran}
\usepackage{cite}
\usepackage{amssymb,amsfonts}
\usepackage{algorithmic}
\usepackage{graphicx}
\usepackage{textcomp}
\usepackage{mathtools}
\usepackage{changes}
\usepackage{subcaption}
\usepackage{algorithm}
\usepackage{xcolor}
\usepackage{color, soul}
\usepackage{amsmath}
\usepackage{booktabs}
\usepackage{multirow}
\usepackage{xstring}
\usepackage{hhline}
\usepackage{siunitx}
\usepackage{physics}
\usepackage{tikz}

\newcommand{\rom}[1]{\uppercase\expandafter{\romannumeral #1\relax}}

\ifCLASSINFOpdf
\else
\fi
\usepackage{url}

\begin{document}
%
\title{ERASOR: Egocentric Ratio of Pseudo Occupancy-based Dynamic Object Removal for Static 3D Point Cloud Map Building}
%
%
%

\author{Hyungtae Lim$^{1}$, \textit{Student Member, IEEE}, Sungwon Hwang$^{2}$, and Hyun Myung$^{1}$, \textit{Senior Member, IEEE}%
\thanks{Manuscript received: October, 15, 2020; Revised January, 10, 2021; Accepted February, 5, 2021.}
\thanks{This paper was recommended for publication by Editor S. Behnke upon evaluation of the Associate Editor and Reviewers' comments.
This work was supported by the Industry Core Technology Development Project, 20005062, Development of Artificial Intelligence Robot Autonomous Navigation Technology for Agile Movement in Crowded Space, funded by the Ministry of Trade, Industry \& Energy (MOTIE, Republic of Korea) and by the research project “Development of A.I. based recognition, judgement and control solution for autonomous vehicle corresponding to atypical driving environment,” which is financed from the Ministry of Science and ICT (Republic of Korea) Contract No. 2019-0-00399. The students are supported by the BK21 FOUR from the Ministry of Education (Republic of Korea).} 
\thanks{$^{1} $H. Lim and H. Myung are with School of Electrical Engineering, KI-AI, KI-R at KAIST (Korea Advanced Institute of Science and Technology), Daejeon, 34141, South Korea {\tt\footnotesize \{shapelim, hmyung\}@kaist.ac.kr}. $^{2} $S. Hwang is with Robotics Program at KAIST, Daejeon, 34141, South Korea {\tt\footnotesize shwang.14@kaist.ac.kr}.}%
\thanks{Digital Object Identifier (DOI): see top of this page.}
}

\maketitle
\IEEEpeerreviewmaketitle


\begin{abstract}
Scan data of urban environments often include representations of dynamic objects, such as vehicles, pedestrians, and so forth. However, when it comes to constructing a 3D point cloud map with sequential accumulations of the scan data, the dynamic objects often leave unwanted traces in the map. These traces of dynamic objects act as obstacles and thus impede mobile vehicles from achieving good localization and navigation performances. To tackle the problem, this paper presents \textcolor{black}{a novel static map building method called \textit{ERASOR}, Egocentric RAtio of pSeudo Occupancy-based dynamic object Removal, which is fast and robust to motion ambiguity. Our approach directs its attention to the nature of most dynamic objects in urban environments being inevitably in contact with the ground.} Accordingly, we propose the novel concept called \textit{pseudo occupancy} to express the occupancy of unit space and then discriminate spaces of varying occupancy. Finally, Region-wise Ground Plane Fitting (R-GPF) is adopted \textcolor{black}{to distinguish static points from dynamic points within the candidate bins that potentially contain dynamic points.} As experimentally verified on SemanticKITTI, our proposed method yields promising performance against state-of-the-art methods overcoming the limitations of existing ray tracing-based and visibility-based methods.
\end{abstract}


%
\IEEEpeerreviewmaketitle

\section{Introduction and Related Works}

\IEEEPARstart{F}{or} most mobile platforms such as Unmanned Ground Vehicles (UGVs), Unmanned Aerial Vehicles (UAVs), or autonomous cars, retrieving \textcolor{black}{a point cloud map} is essential in achieving long-term autonomy\textcolor{black}{\cite{lim2020normal,kim2019gp,kim20191daylearning,behley2018efficient, behley2019semantickitti,kim2018scancontext,kim2017robust,pomerleau2014long, hawes2017strands,banerjee2019lifelong}.} Robots may utilize static information of \textcolor{black}{their} neighboring geometrical structure provided by \textcolor{black}{the map} to achieve successful localization or navigation. \textcolor{black}{
The} map can be represented in many different forms, such as feature map-based representation \cite{sarlin2019coarse}, occupancy-based representation \cite{banerjee2019lifelong, wurm2010octomap}, or structural-variance-robust place representation \cite{kim20191daylearning}. This paper limits the scope of the map to a 3D point cloud map, which is generated by accumulations of laser scans obtained by \textcolor{black}{a} three-dimensional LiDAR sensor.

\begin{figure}[h]
	\centering
	\begin{subfigure}[b]{0.23\textwidth}
		\includegraphics[width=1.0\textwidth]{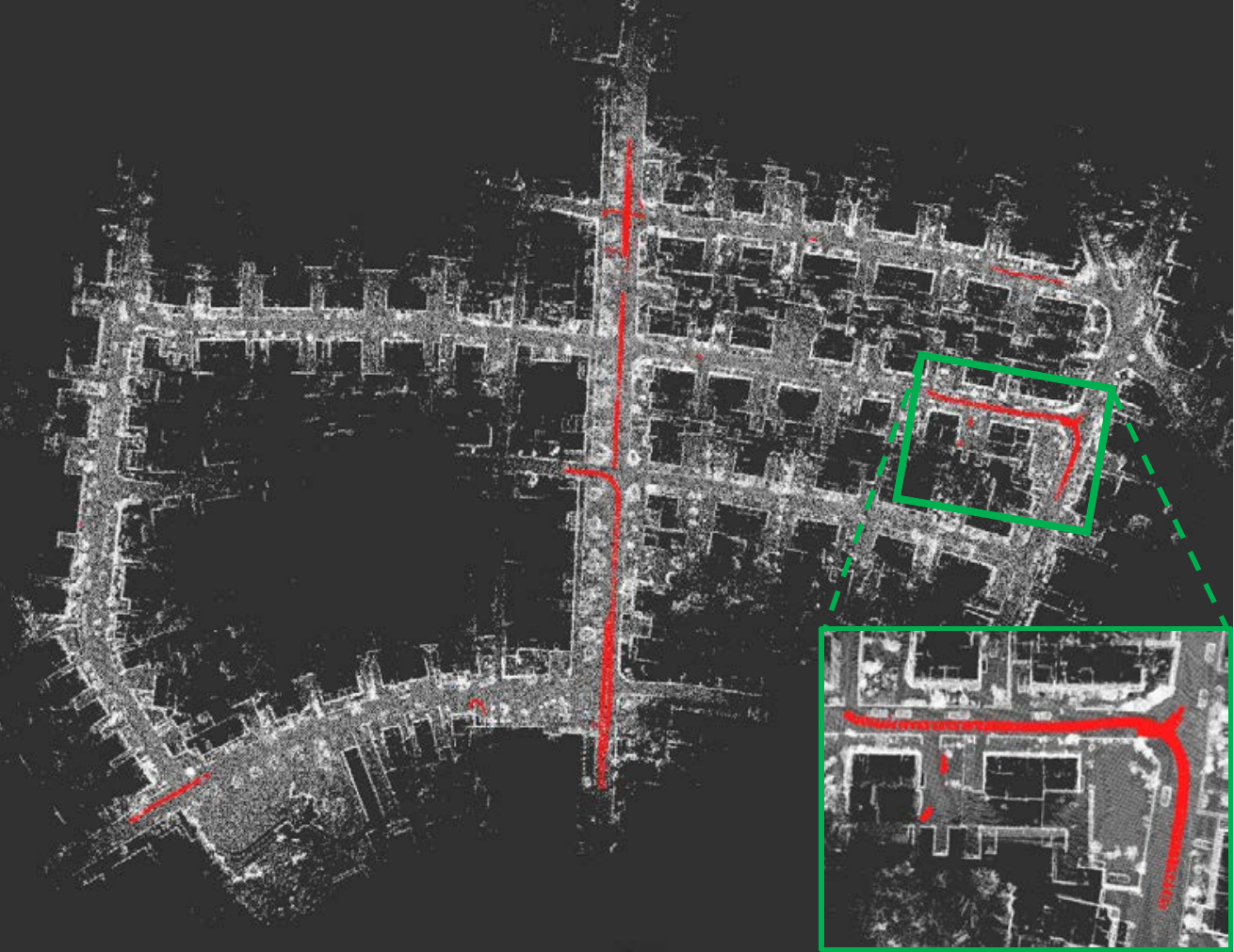}
		\label{fig:seq_05_before}
	\end{subfigure}
	\begin{subfigure}[b]{0.23\textwidth}
		\includegraphics[width=1.0\textwidth]{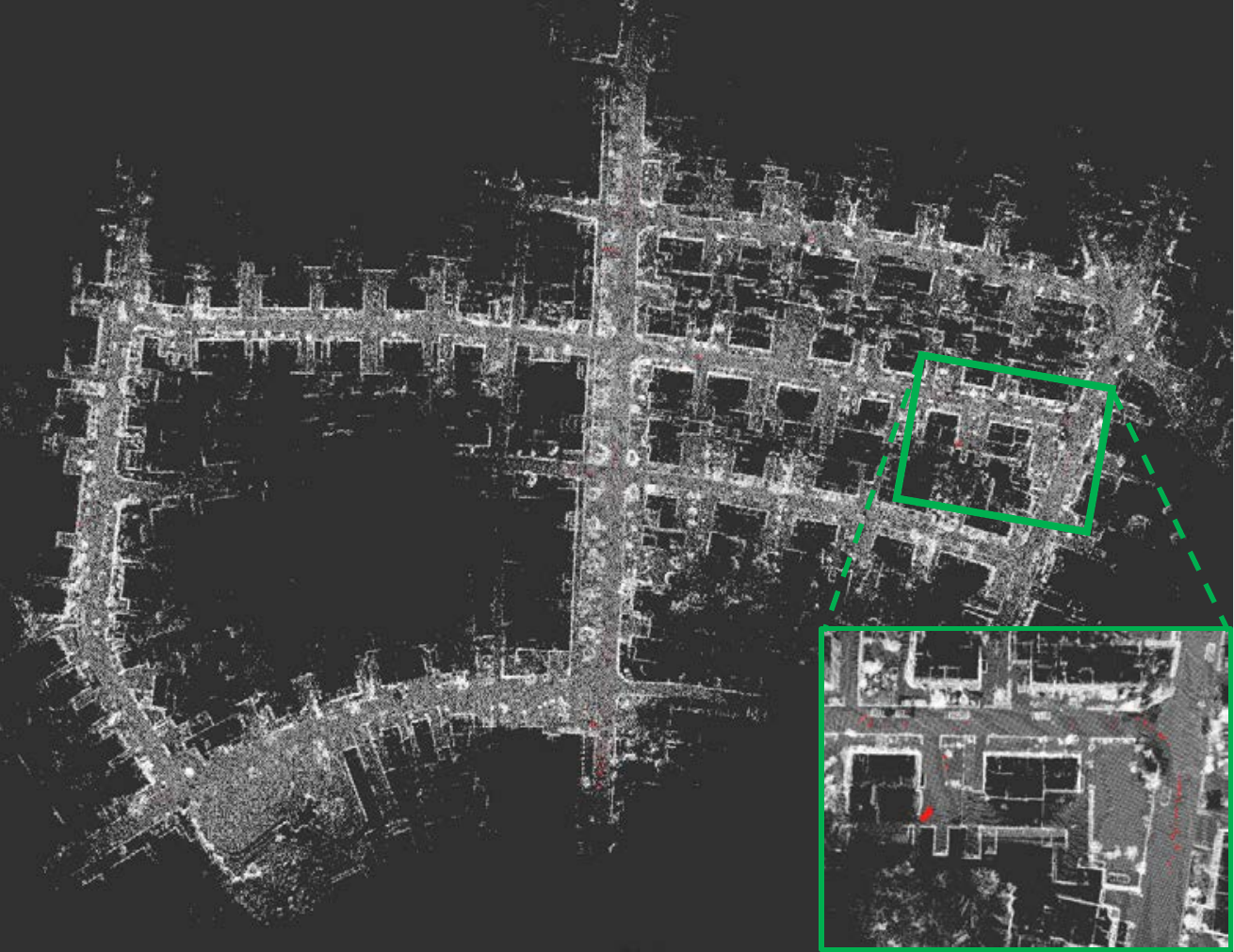}
		\label{fig:seq_05_after}
	\end{subfigure}
	\caption{\textcolor{black}{Before and after the application of our proposed method called \textit{ERASOR} on sequence \texttt{05} of SemanticKITTI. Dynamic objects are highlighted in red (best viewed in color).}}
	\label{fig:seq_05_whole}
\end{figure}

Unfortunately, because scan data presents a snapshot of the surroundings, scan data in urban environments inevitably includes representations of dynamic objects, such as vehicles, pedestrians, and so forth \cite{pagad2020robust, schauer2018peopleremover,kimremove,yoon2019mapless}. Moreover, because a 3D point cloud map is the product of sequential accumulations of the scan data, there \textcolor{black}{might} be traces of dynamic objects, or the \textit{ghost trail effect} \cite{pagad2020robust,pomerleau2014long}, as shown in Fig.~\ref{fig:seq_05_whole} on the left. These traces of dynamic objects act as obstacles in the map and thus impede mobile robots from exhibiting good localization and navigation performance. 

Conventional methods that tackle this problem are mainly divided into two categories: a) rejection of dynamic objects at the time of map generation \cite{yoon2019mapless,elfes1989occugrid} and b) post-rejection of dynamic objects with a complete set of generated map \cite{pomerleau2014long,kimremove,kim2017robust}. The methodology presented in this paper focuses on the latter case. Accordingly, the latter category can be further classified into three main methodologies: a) segmentation-based, b) ray tracing-based, and c) visibility-based methods.


\textcolor{black}{First of all, segmentation-based methods are usually based on clustering \cite{yoon2019mapless,litomisky2012removing, yin2015removing}. Litomisky and Bhanu \cite{litomisky2012removing} discriminated dynamic clusters among static clusters by using the Viewpoint Feature Histogram (VFH) and Yin \textit{et al.} \cite{yin2015removing} utilized feature matching to check false correspondences and then used them as seeds to extract clusters from dynamic objects. Yoon \textit{et al.} \cite{yoon2019mapless} proposed a region growing-based method.}

In addition, the deep learning-based semantic segmentation or detection has led to additional significant improvement in performance\cite{milioto2019rangenet++}. Given correct segment\textcolor{black}{ed} points with dynamic labels, a map can be straightforwardly constructed by excluding the predicted dynamic points \cite{sun2018recurrent,kimremove}. However, existing segmentation-based approaches only focus on the dynamic points rejection at the moment and not show the final static map. Besides, these are heavily rely on the supervised labels; therefore, they are vulnerable to label\textcolor{black}{ing} errors or dynamic objects of unlabeled classes \cite{wong2020identifying}. To mitigate this problem, considering the scan-to-map relationship is necessary.

\textit{Occupancy grid} \cite{elfes1989occugrid} and \textit{Octomap} \cite{wurm2010octomap} are typical methods that use ray tracing, which count hits and misses of scans in the \textcolor{black}{gridmap} space and decide the occupancy of the space. By extension, Schauer and N{\"u}chter \cite{schauer2018peopleremover} proposed the removal of dynamic points by traversing a voxel occupancy grid and Pagad \textit{et al.} \cite{pagad2020robust} suggested a combinination of object detection and Octomap. However, ray tracing-based methods are computationally expensive, \textcolor{black}{which led to the introduction of visibility-based methods} to reduce the computational cost \cite{kimremove, pomerleau2014long, banerjee2019lifelong}.

\begin{figure}[h]
	\centering
	\begin{subfigure}[b]{0.48\textwidth}
		\includegraphics[width=1.0\textwidth]{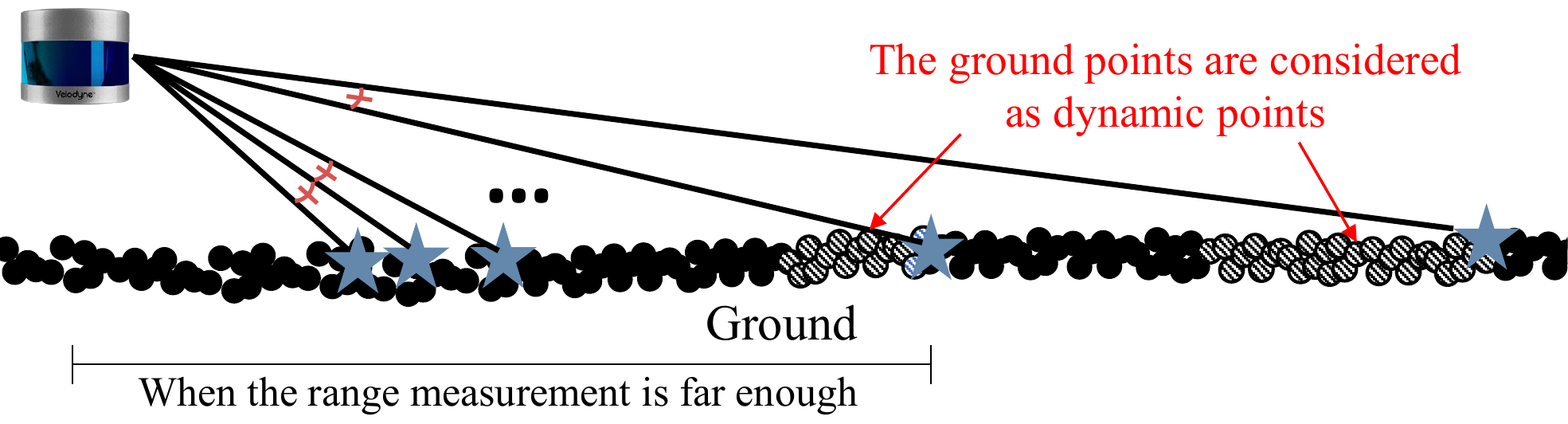}
		\caption{}
	\end{subfigure}
	\begin{subfigure}[b]{0.23\textwidth}
		\includegraphics[width=1.0\textwidth]{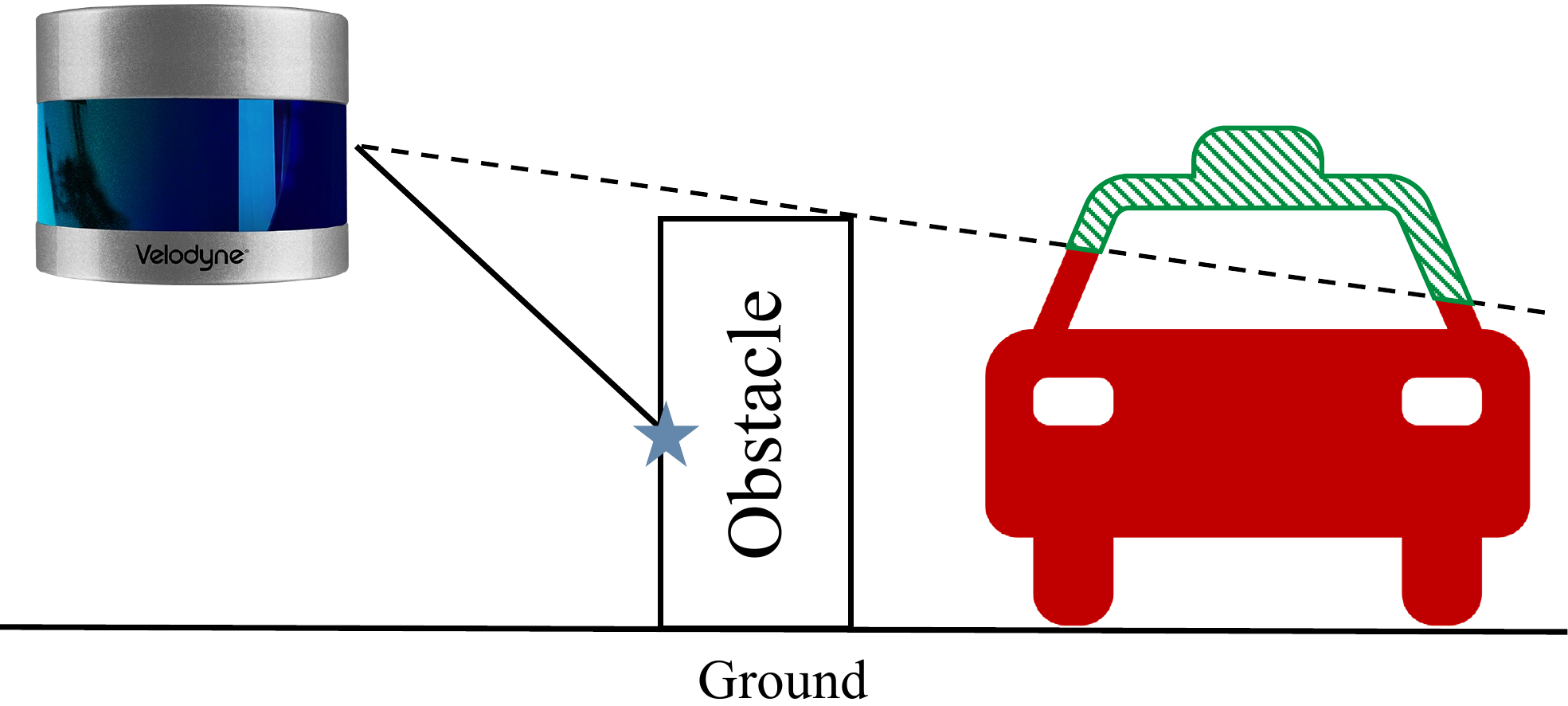}
		\caption{}
	\end{subfigure}
	\begin{subfigure}[b]{0.23\textwidth}
		\includegraphics[width=1.0\textwidth]{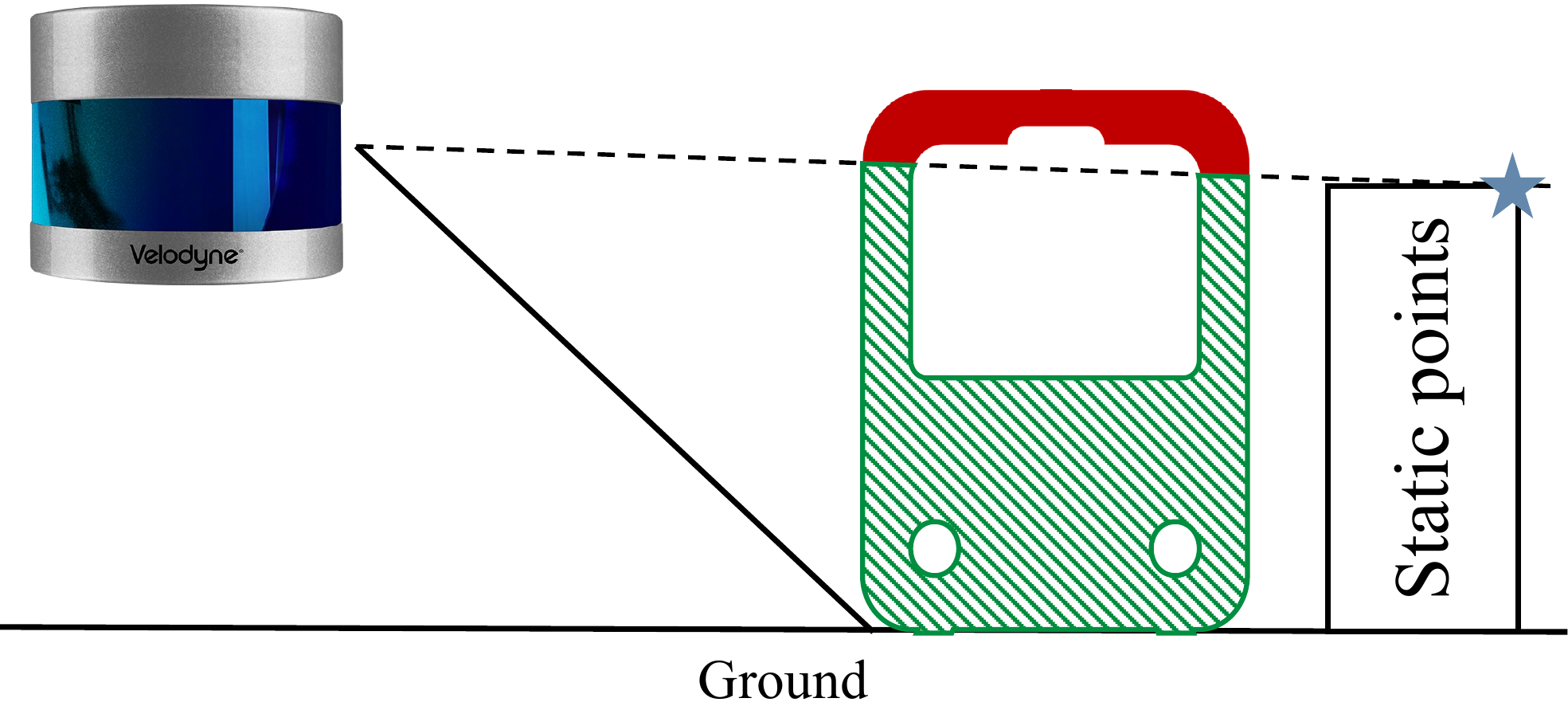}
		\caption{}
	\end{subfigure}
	\caption{Potential limitations of the ray tracing-based methods and visibility-based methods. (a) \textcolor{black}{I}ncidence angle ambiguity (b) \textcolor{black}{O}cclusion (c) \textcolor{black}{I}rremovable dynamic points when there are no static points behind them. The highlighted parts in green are erasable by previous methods, and parts in red are not (best viewed in color).}
	\label{fig:visibility_limitation}
\end{figure}

Visibility-based methods are based on the premise that \textcolor{black}{if a query point is observed behind a previously acquired point in the map, then the previously acquired point is dynamic \cite{yoon2019mapless,kimremove, pomerleau2014long}}. However, this assumption has \textit{incidence angle ambiguity}; as the range measurement from the ground becomes longer, the incidence angle of a point becomes more ambiguous, so the neighboring ground points may be falsely considered as dynamic objects (Fig.~\ref{fig:visibility_limitation}(a)). To resolve the problem, Pomerleau \textit{et al.} \cite{pomerleau2014long} utilized \textcolor{black}{a} Bayesian approach using normal and incidence angles to represent \textcolor{black}{the} state for each point. By extension, Kim and Kim \cite{kimremove} proposed a pixel-to-window comparison method to take incidence angle ambiguity into account\textcolor{black}{. 
However,} all previous works struggled with the occlusion issue (Fig.~\ref{fig:visibility_limitation}(b)) and it was not possible to remove the dynamic points of huge dynamic objects once static points are not \textcolor{black}{observed} behind the dynamic points (Fig.~\ref{fig:visibility_limitation}(c)).

In this paper, we propose a novel static map building method, referred to as \textcolor{black}{Egocentric RAtio of pSeudo Occupancy-based dynamic object Removal (ERASOR)}, to overcome the limitations of the aforementioned methods. Our method is a visibility-free approach and leverages our proposed representation of points organized in vertical columns called \textit{pseudo occupancy}. The contribution of this paper is fourfold:
\begin{itemize}
	\item We propose a fast and robust method called \textit{Scan Ratio Test} (SRT) \textcolor{black}{to fetch bins which contain dynamic points based on the nature of most dynamic objects in urban environments, such as terrestrial vehicles and pedestrians, being inevitably in contact with the ground.}
	\item After fetching the bins, we apply \textit{Region-wise Ground Plane Fitting} (R-GPF), \textcolor{black}{a novel static point retrieval method with low computational load. R-GPF overcomes the potential limitations of ray tracing-based methods and visibility-based methods, which are presented in Fig.~\ref{fig:visibility_limitation} with details.}
	\item Validations against other state-of-the-art methods are conducted. \textcolor{black}{To the best of our knowledge, it is the first attempt to conduct quantitative comparisons with other state-of-the-art methods on SemanticKITTI \cite{behley2019semantickitti}.} In particular, we point out the limitation of precision/recall metrics and present alternative metrics suitable for static map building tasks: \textit{Preservation Rate} and \textit{Rejection Rate}.
	\item Our proposed method shows a promising performance over state-of-the-art methods. \textcolor{black}{In particular, ERASOR removes dynamic points with the least loss of static points and it is at least ten times faster than state-of-the-art methods.}
\end{itemize}

The rest of the paper is organized as follows: Section~\rom{2} explains the concept of pseudo occupancy and our proposed method. Section~\rom{3} describes the experiments and novel error metrics, and the experimental results \textcolor{black}{are discussed in Section~\rom{4}}. Finally, Section~\rom{5} summarizes our contributions and describes future works.

\section{Static Map Building via ERASOR}


\begin{figure*}[ht]
    \centering
    \begin{subfigure}[b]{.97\linewidth}
        \includegraphics[width=\linewidth]{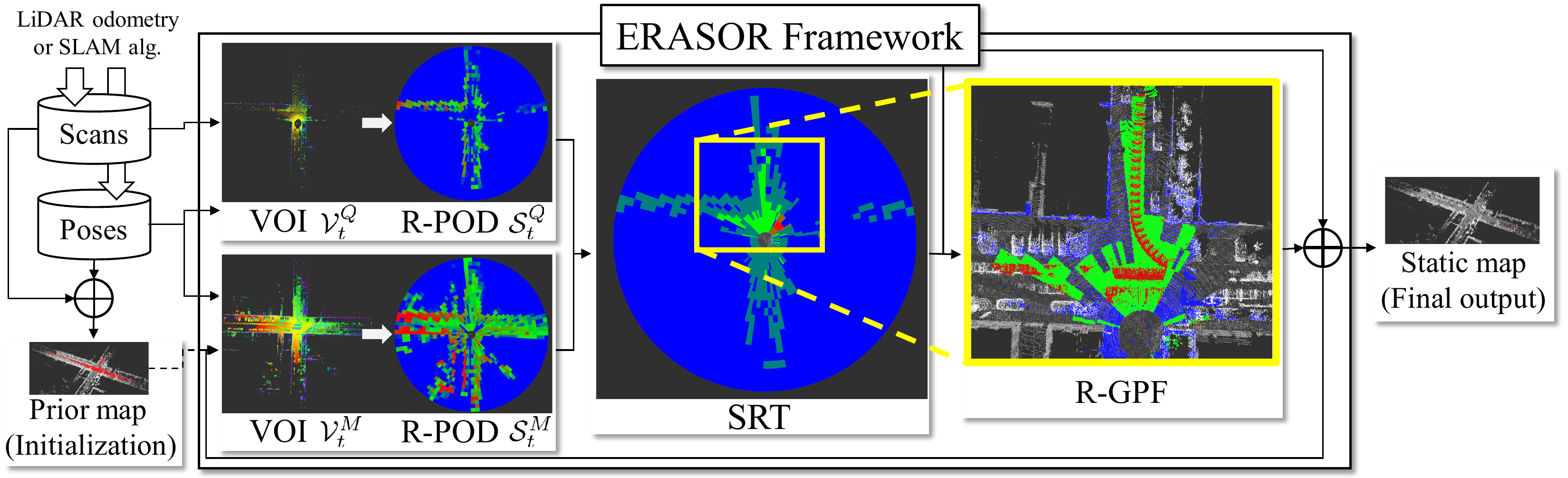}
    \end{subfigure}
    \caption{\textcolor{black}{Overview of our proposed method. A query cloud and a map cloud are each encoded into the \textit{Volume of Interest} (VOI) followed by the \textit{Region-wise Pseudo Occupancy Descriptor} (R-POD) to express the occupancies of unit spaces using \textit{pseudo occupancy}. Then, the \textit{Scan Ratio Test} (SRT) takes R-PODs from each cloud to discriminate spaces of varying occupancy. After fetching bins that potentially contain dynamic points (\textcolor{green}{green} on SRT), \textit{Region-wise Ground Plane Fitting} (R-GPF) discriminates ground points, which are reverted into the map (\textcolor{green}{green} on R-GPF) while dynamic points above the reverted grounds are rejected (\textcolor{red}{red} on R-GPF). Best viewed in color.}}
    
    \label{fig:overview_erasor}
\end{figure*}

The schematic diagram of our proposed method on static map building is illustrated in Fig.~\ref{fig:overview_erasor}. The following paragraphs highlight the problem definition and the reasoning behind each module of ERASOR. Note that this paper mainly focuses on instance-level dynamic object rejection for static map building. That is, large-scale changes in urban environments, such as the redevelopment or restoration of buildings, are beyond the scope of this work.

\subsection{Problem Definition} 

Let $\mathcal{P}_t=\{\mathbf{p}_{1}, \mathbf{p}_{2}, \dots, \mathbf{p}_{n}\}$ be a set of point cloud that contains $n$ points \textcolor{black}{at} time step $t$, where each point is expressed as $\mathbf{p}_{k}=\{x_k, y_k, z_k\}$ in Cartesian coordinates. \textcolor{black}{Let $\mathcal{P}^{Q}_{t}$ be the query point cloud on the query frame and $^{W}_{Q}\mathbf{T}_{t}$ be SE(3) pose associated with $\mathcal{P}^{Q}_{t}$. In this paper, \textcolor{black}{it is assumed that the set of poses are given after optimization or correction through registration. By letting $^{W}_{Q}\mathbf{T}_{t} * \mathcal{P}^{Q}_{t}$ be transformed representation of $\mathcal{P}^{Q}_{t}$ in the world frame $W$,} the prior map built with a set of raw LiDAR scans, $\mathcal{M}$, can be formulated as follows:}
\begin{equation}
\textcolor{black}{\mathcal{M}=\bigcup_{t\in\left[T\right]}{^{W}_{Q}\mathbf{T}_{t} * \mathcal{P}^{Q}_{t}}}
\end{equation}  
where $T$ is the total time step and $[T]$ is equal to $\{1, 2, \dots, T-1, T\}$. \textcolor{black}{Note that $\mathcal{M}$ is on the world frame and contains all measured dynamic points.} 

\textcolor{black}{Next, let} $\mathcal{P}^{M}_{t}$ be the submap of $\mathcal{M}$ represented with respect to the query point cloud coordinate, \textcolor{black}{which means the submap of $\mathcal{M}$ is transformed via $^{Q}_{W}\mathbf{T}_{t}={^{W}_{Q}\mathbf{T}_{t}}^{-1}$.} Throughout this paper, the superscripts $Q$ and $M$ are used to represent the query and map, respectively. By letting $\hat{\mathcal{M}}$ be the estimated static map, the problem of our interest is defined as follows:
\begin{equation}
\hat{\mathcal{M}}=\mathcal{M}-\bigcup_{t\in\left[T\right]}{\hat{\mathcal{M}}_{dyn, t}} 
\end{equation}  
where $\mathcal{\hat{\mathcal{M}}}_{dyn, t}$ refers to the estimated dynamic points determined \textcolor{black}{by} discrepancies between $\mathcal{P}^{Q}_{t}$ and $\mathcal{P}^{M}_{t}$.

Before the introduction of our visibility-free method, explanations of the aspects of a unit space must be given. Our approach directs its attention to the nature of most dynamic objects in urban environments, such as terrestrial vehicles and pedestrians, being inevitably in contact with the ground. Based on this assumption, we may construct four possible cases: 

\begin{itemize} 
	\item An object is on the \textcolor{black}{ground} in $\mathcal{P}^M_{t}$, while the \textcolor{black}{ground} of the same position is free from an object in $\mathcal{P}^Q_{t}$.
	\item The \textcolor{black}{ground} is free from an object in $\mathcal{P}^M_{t}$, while there is an object on the \textcolor{black}{ground} of the same position in $\mathcal{P}^Q_{t}$.
	\item An object is on the \textcolor{black}{ground} in both $\mathcal{P}^M_{t}$ and $\mathcal{P}^Q_{t}$.
	\item There are no objects on the \textcolor{black}{ground} in $\mathcal{P}^M_{t}$ and $\mathcal{P}^Q_{t}$.
\end{itemize}

Of these cases, points that represent dynamic objects will be categorized to \textcolor{black}{the first case and the second case. However, our goal is to refine the map cloud, so the second case is neglected. For clarity, we define two terms; \textit{potentially dynamic case} for the first case and \textit{definitely static case} for the third and fourth cases.}

Thus, our objective can be paraphrased to detect the unit spaces in $\mathcal{P}^M_{t}$ that satisfy \textcolor{black}{potentially dynamic case}, and to retrieve regions that include dynamic points (see Section~\rom{2}.\textit{D}). The set of unit spaces can then be refined by removing the dynamic parts \textcolor{black}{(see Section~\rom{2}.\textit{E})}.

\subsection{Volume of Interest} 
Before $\mathcal{P}^M_{t}$ and $\mathcal{P}^Q_{t}$ are separated into unit spaces, physically significant domain of point space, or \textit{Volume of Interest} (VOI, $\mathcal{V}_{t}$), is defined and formulated as follows: 

\begin{equation}
\mathcal{V}_{t}=\{\mathbf{p}_{k}  \mid \mathbf{p}_{k} \in \mathcal{P}_{t},\; \rho_{k}<L_{\max},\; h_{\min}<z_k< h_{\max}\}
\end{equation}
where $\rho_{k}=\sqrt{x_{k}^{2}+y_{k}^{2}}$; $L_{\max}$, $h_{\min}$, and $h_{\max}$ are \textcolor{black}{constant parameters} that indicate the maximum radial boundary, the minimum height of the VOI, \textcolor{black}{and the maximum height of the VOI} with respect to the ground, respectively. Heights of dynamic objects of our interest, such as vehicles or pedestrians, usually fall into a reasonable range \cite{kim2017robust, kim2018scancontext}. Therefore, we may set \textcolor{black}{$h_{\max}=3.0$m, $h_{\min}=-1.0$m and $L_{\max}=80.0$m}. Note that the minimum value \textcolor{black}{$h_{\min}=-1.0$m}, which is lower than the ground, i.e. \textcolor{black}{$h=0$m}, is set to cover sloped areas as well as the uncertainty of the detected height of the ground. 

\textcolor{black}{Note that $\mathcal{V}^{M}_t$ is extracted from $\mathcal{M}$ by neighbor search, i.e. K-D Tree \cite{bentley1975multidimensional}, with respect to the position of the query frame followed by transformation via ${^{W}_{Q}\mathbf{T}_{t}}^{-1}$ to reduce computational cost. This is because $\mathcal{M}$ mostly consists of more than a million points; thus, it is not cost-effective to transform all points to the query frame and then extract $\mathcal{V}^{M}_t$.}

\subsection{Region-wise Pseudo Occupancy Descriptor} \label{sec:src}
For region-wise dynamic object removal, the VOIs are encapsulated by the egocentric spatial occupancy descriptor called \textit{Region-wise Pseudo Occupancy Descriptor} (R-POD). Its design is motivated by \textit{Cells of Interest} (COI) \cite{kim2017robust} and \textit{Scan Context} \cite{kim2018scancontext}.

\textcolor{black}{Kim \textit{et al.} \cite{kim2017robust}} proposed COI to encode vertical information in a point cloud into binary bits. However, COI uses RGB format and is not LiDAR-friendly \cite{kim20191daylearning}. On the other hand, \textcolor{black}{Scan Context \cite{kim2018scancontext,wang2020intensity}} takes the maximum height from each bin and arranges them into a 2D matrix in an egocentric way. It is efficient to check the scan-to-scan relationship, yet it may be risky to encode the absolute maximum of z-directional information to check the scan-to-map relationship. That is because z-directional information of poses from LiDAR \textcolor{black}{
o}dometry or LiDAR SLAM is usually more uncertain than x and y directional information \cite{zhang2014loam}. 

To this end, our R-POD blends the best of both methods by defining a vertical bin that expresses occupancy with the boundary difference of vertical information, or \textit{pseudo occupancy}, in an egocentric way. Similar to Scan Context, R-POD takes $\mathcal{V}_{t}$ and divides the volume over the regular interval of azimuthal and radial directions, i.e. \textit{sectors} and \textit{rings}. Let $N_{r}$ and $N_{\theta}$ be the numbers of rings and sectors. Then R-POD, which is denoted as \textcolor{black}{$\mathcal{S}_{t}$}, can be represented as follows:
\begin{equation}
\mathcal{S}_{t}=\bigcup_{i\in\left[N_{r}\right], j\in\left[N_{\theta}\right]} \mathcal{S}_{(i,j), t}
\end{equation}
where \textcolor{black}{$\mathcal{S}_{(i,j), t}$} denotes the $(i,j)$-th bin of R-POD \textcolor{black}{at} time step $t$. Let \textcolor{black}{$\theta=\arctan2(y, x)$}. Then, each $\mathcal{S}_{(i,j), t}$ consists of the cloud points that satisfy the following condition:
\textcolor{black}{
\footnotesize{
\begin{equation}
\begin{aligned}
\mathcal{S}_{(i,j), t}=\left\{\mathbf{p}_{k} \mid \mathbf{p}_{k} \in \mathcal{V}_{t},\; \frac{(i-1) \cdot L_{\max }}{N_{r}} \leq \rho_{k}<\frac{i \cdot L_{\max }}{N_{r}},\right.\\
\left.\frac{(j-1) \cdot 2 \pi}{N_{\theta}}-\pi \leq \theta_{k}<\frac{j \cdot 2 \pi}{N_{\theta}}-\pi\right.\}.
\end{aligned}
\end{equation}}}

Thereafter, the unit space, i.e. each bin, assigns a single real value to describe pseudo occupancy, $\Delta h_{(i,j), t}$. Let \textcolor{black}{$Z_{(i,j), t}=\{z_k\in\mathbf{p}_{k}|\mathbf{p}_{k}\in\mathcal{S}_{(i,j), t}\}$.} Then, \textcolor{black}{the} pseudo occupancy of each bin is \textcolor{black}{encoded} as follows: 
\textcolor{black}{
\begin{equation}
\Delta h_{(i,j), t} = \sup{\{Z_{(i,j), t}\}} - \inf{\{Z_{(i,j), t}\}}
\end{equation}}where $\sup$ and $\inf$ denote supremum and infimum, respectively.

\subsection{Scan Ratio Test} \label{sec:srt}
\textit{Scan Ratio Test} (SRT) is proposed to \textcolor{black}{check whether discrepancy of pseudo occupancy occur} given a pair of R-PODs of query and map point clouds, i.e. $\mathcal{S}^{Q}_{t}$ and $\mathcal{S}^{M}_{t}$. The SRT is motivated by Lowe's \textit{Ratio Test} in Scale Invariant Feature Transform (SIFT) \cite{lowe2004distinctive}. The ratio-based approach is shown to be more robust against the change of scenes over the global threshold-based method\cite{lowe2004distinctive}. Due to this advantage and its generality, we may apply the concept of the ratio test to our methodology when comparing pseudo occupancy. 

\begin{figure}[h]
    \centering
    \begin{subfigure}[b]{.46\linewidth}
        \includegraphics[width=\linewidth]{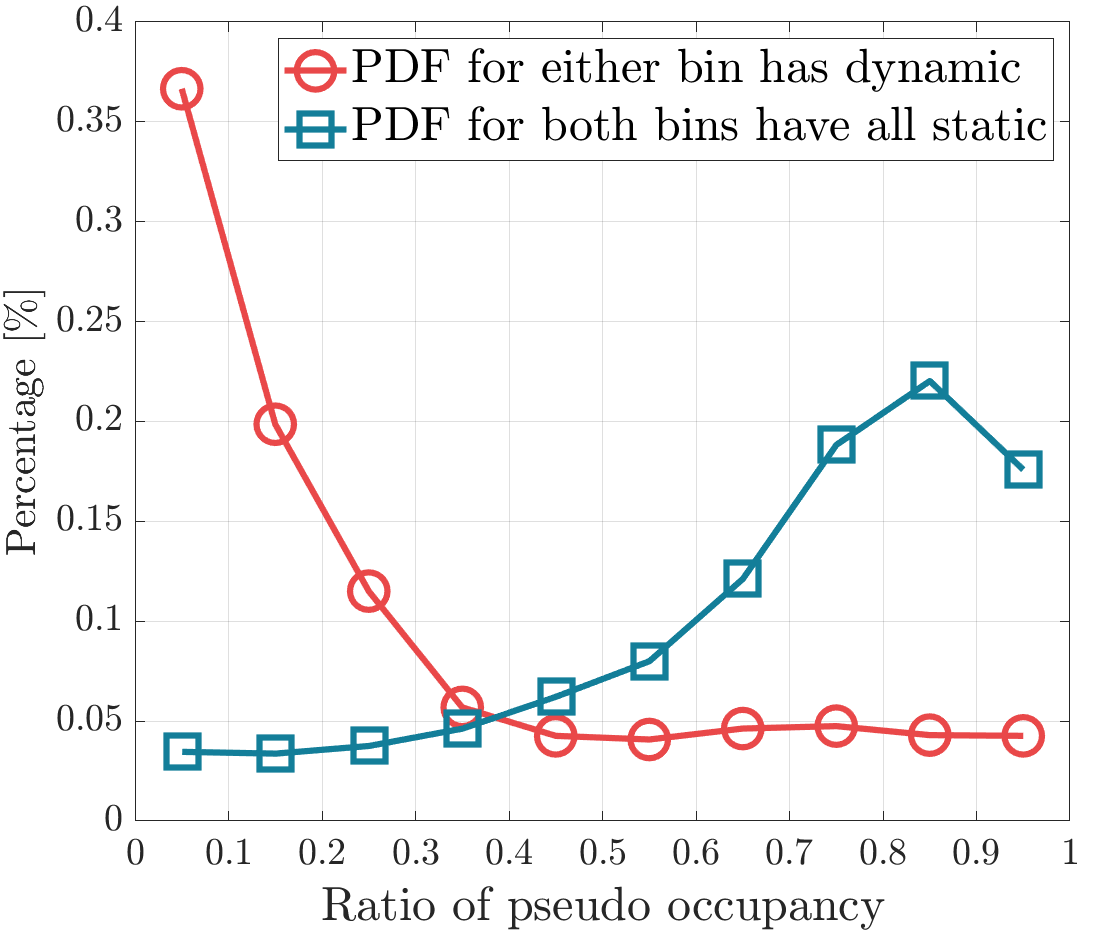}
        \caption{}
    \end{subfigure}
    \begin{subfigure}[b]{.46\linewidth}
        \includegraphics[width=\linewidth]{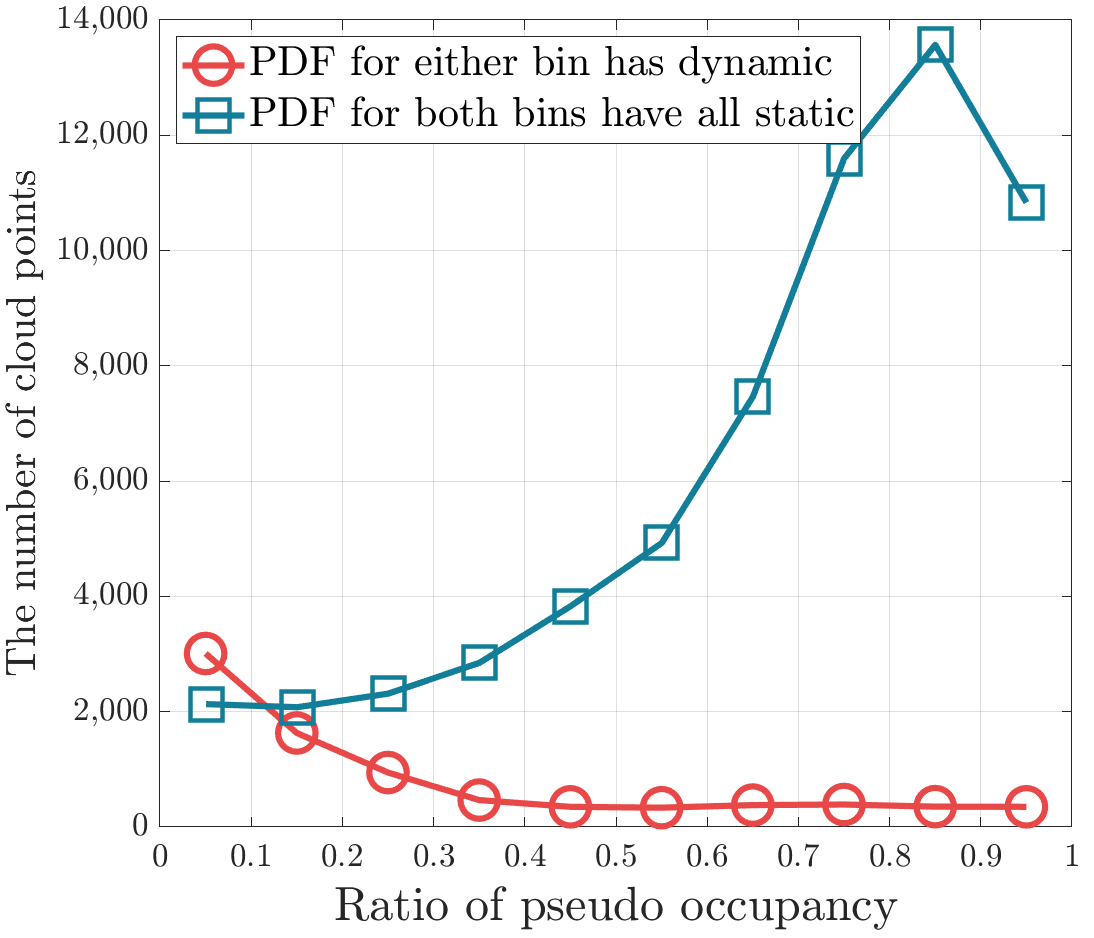}
        \caption{}
    \end{subfigure}
    \caption{(a) Probability Distribution Function (PDF) of the ratio of pseudo occupancy between two corresponding potentially dynamic bins \textcolor{black}{(b) Histogram of the actual number of cloud points between two corresponding potentially dynamic bins.}}
    \label{fig:pdf}
\end{figure}

Let the term \textit{scan ratio} be the ratio of pseudo occupancy between each pair of bins in $\mathcal{S}^{Q}_{t}$ and $\mathcal{S}^{M}_{t}$, i.e. the ratio between $\Delta h_{(i,j), t}^Q$ and $\Delta h_{(i,j), t}^M$. The scan ratio is then used to categorize bins to the aforementioned cases and to select bins that belong to \textcolor{black}{the potentially dynamic case where either bin includes dynamic points of our interest.} Straightforwardly, the scan ratio must be close to 1 if there are no changes in both bins, \textcolor{black}{i.e. the definitely static case, whereas the scan ratio that belongs to the potentially dynamic case} must be much smaller than 1, which is caused by the presence of a dynamic object \textcolor{black}{on the ground} as mentioned in \textcolor{black}{Section~\rom{2}.\textit{A}}.

Therefore, we select bins with a scan ratio smaller than a ratio threshold as bins that may include dynamic objects, or \textit{potentially dynamic bins}. Fig.~\ref{fig:pdf}(a) illustrates the Probability Distribution Function (PDF) of the scan ratio and \textcolor{black}{Fig.~\ref{fig:pdf}(b) describes the histogram of the number of cloud points using the SemanticKITTI dataset} (see Section~\rom{3}.\textit{A}). A set of bins including dynamic points either in the query or in the map tends to have a small scan ratio. In contrast, a set of bins with only static points shows a scan ratio close to 1. 

\textcolor{black}{Based on these observations, the ratio threshold is set to 0.2, which \textcolor{black}{is} empirically determined to be strict enough. Fig.~\ref{fig:overview_erasor} illustrates the result retrieved from SRT. The green bins on SRT are potentially dynamic bins, and the teal bins are those whose scan ratio is over 0.2. The red bins are deterministically static regions or already cleaned regions, i.e. $\Delta h_{(i,j), t}^M / \Delta h_{(i,j), t}^Q <0.2$. Note that if either the map or query bin contains few points, SRT is skipped, where those are colored in blue on the SRT in Fig.~\ref{fig:overview_erasor}.}


\subsection{Region-wise Ground Plane Fitting}
 
 
According to the assumption in \textcolor{black}{Section~\rom{2}.\textit{A}}, it is observed that many potentially dynamic bins solely consist of ground points and dynamic points. Therefore, we may think of the low-computationally demanding static point retrieval method.

Bin-wise estimation of the ground plane has advantages over the estimation of a single plane model of the whole map, because many urban areas include grounds with curbs or sloped regions that do not form perfect planes \cite{zermas2017fast}. \textcolor{black}{For this reason, bin-wise ground plane fitting is adopted, which is motivated by bin-wise line fitting to find ground points \cite{himmelsbach2010fast}.} Because each bin is small relative to the whole map, it is safe to assume that the ground is planar within the bin. As a result, R-GPF can robustly select and revert more static points back to the map on the non-planar, urban specific regions (see \textcolor{black}{Section~\rom{4}.\textit{B}}).

Let the $l$-th bin, i.e. $\mathcal{S}^{M}_{l, t}$, be the candidate where the total $L$ potentially dynamic bins are selected among $N_{r} \times N_{\theta}$ bins throughout the SRT. Because the points of each bin with the lowest heights are most likely to belong to the ground surface \cite{himmelsbach2010fast}, the lowest-height seed points are selected from $\mathcal{S}^{M}_{l, t}$, and let $\bar{z}$ be the mean z value of the selected seed points. \textcolor{black}{Then, the initial estimated ground points set $^{0}\mathcal{I}_{l,t}$ is obtained as follows:}

\begin{equation}
\begin{aligned}
\textcolor{black}{^{0}\mathcal{I}_{l,t}=\left\{\mathbf{p}_{k} \mid \mathbf{p}_{k} \in \mathcal{S}^{M}_{l, t}, \; z(\mathbf{p}_{k}) < \bar{z} + \tau_{\text{seed}} \right.\}}
\end{aligned}
\end{equation}
\textcolor{black}{where $z(\cdot)$ represents the z value of a point and $\tau_{\text{seed}}$ denotes height margin. Because our method is iterative, let the $i$-th inliers be $^{i}\mathcal{I}_{l,t}$, then the covariance matrix $^iC_{l,t}$ of $^{i}\mathcal{I}_{l,t}$ is calculated as follows:}

\begin{equation}
\textcolor{black}{^iC_{l,t}=\sum_{j=1:|^{i}\mathcal{I}_{l,t}|}\left(\mathbf{p}_{j}-^i\bar{\mathbf{p}}_{l,t}\right)\left(\mathbf{p}_{j}-^i\bar{\mathbf{p}}_{l,t}\right)^{T}}
\end{equation}
\textcolor{black}{where $|\cdot|$ and $^i\bar{\mathbf{p}}_{l,t}$ denote the size of a set and mean position of $^{i}\mathcal{I}_{l,t}$, respectively.}

\textcolor{black}{Next, using Principal Component Analysis (PCA), three eigenvalues and the corresponding three eigenvectors are calculated, i.e. $^iC_{l,t} \vec{v}_m=\lambda_m \vec{v}_m$ where $m=1,2,3$. Then, the eigenvector with the smallest eigenvalue is the most likely to represent the normal vector to the ground.} 
\textcolor{black}{Let the eigenvector be $^i\mathbf{n}_{l,t} = [{^{i}a}_{l, t} \ {^{i}b}_{l, t} \ {^{i}c}_{l, t}]^T$.} Then, the plane coefficient can be calculated as $^i{d}_{l,t}=-^i\mathbf{n}_{l,t}^{T}  {^i\bar{\mathbf{p}}_{l,t}}$, whose plane equation is ${^{i}a}_{l,t} x+{^{i}b}_{l,t} y+{^{i}c}_{l,t} z+{^{i}d}_{l,t}=0$. Finally, our goal is to extract the potentially static points below the plane as follows:

\begin{equation}
\begin{aligned}
\textcolor{black}{^{i+1}\mathcal{I}_{l,t}=\left\{\mathbf{p}_{k} \mid \mathbf{p}_{k} \in \mathcal{S}^{M}_{l, t}, \; {^{i}d}_{l,t} - {^{i}\hat{d}}_k < \tau_{g} \right.\}}
\end{aligned}
\end{equation}
\textcolor{black}{where $^{i}\hat{d}_k=-^{i}\mathbf{n}_{l,t}^{T} \mathbf{p}_k$ and $\tau_{g}$ denotes the distance margin of the plane. The procedure is repeated three times, after which the final $^{3}\mathcal{I}_{l,t}$ is reverted back into the map rejecting dynamic points. Therefore, the dynamic points rejected from our method can be directly retrieved as follows:}

\begin{equation}
\textcolor{black}{\hat{\mathcal{M}}_{dyn, t}=\bigcup_{l\in\left[L\right]}{(\mathcal{S}^{M}_{l, t} - ^{3}\mathcal{I}_{l,t})}.} \end{equation}  

\textcolor{black}{Fig. \ref{fig:overview_erasor} presents the procedure of R-GPF briefly. The red points on R-GPF represent dynamic points and the green points represent the ground points that are reverted back into the map cloud.}

\section{Experiments}

\subsection{Dataset}


For thorough validation, we conducted experiments using the SemanticKITTI dataset \cite{behley2019semantickitti,geiger2012we}. Because SemanticKITTI provides point-wise annotations, points annotated with selected classes (252, 253, 254, 255, 256, 257, and 259) are labeled as ground-truth dynamic points to be removed.

Next, we manually selected the top-5 time frames among the Semantic KITTI dataset that contain the largest number of appearances of dynamic objects to evaluate algorithms quantitatively. \textcolor{black}{Of course, our proposed method is applicable to the whole map as shown in Fig. \ref{fig:seq_05_whole}. However, in general, the number of static points is too much larger than those of dynamic points on the map, and most of the frames contain no or few dynamic points; thus, it is not distinguishable at a glance whether the dynamic points have been removed well or not.}

Therefore, frames \texttt{00} (4,390 - 4,530), \texttt{01} (150 - 250), \texttt{02} (860 - 950), \texttt{05} (2,350 - 2,670), and \texttt{07} (630 - 820) were chosen as our static map construction benchmark where the \textcolor{black}{numbers} in parenthesis indicate \textcolor{black}{the} start and end frames. The selected frames include various environments such as countryside, highway, and intersections. Maps are constructed at the regular intervals with the poses provided by SuMa \cite{behley2018efficient} which contains inherent uncertainty. 

		

\subsection{Error Metrics} 

\begin{figure}[ht]
    \centering
    \begin{subfigure}[b]{.95\linewidth}
        \includegraphics[width=\linewidth]{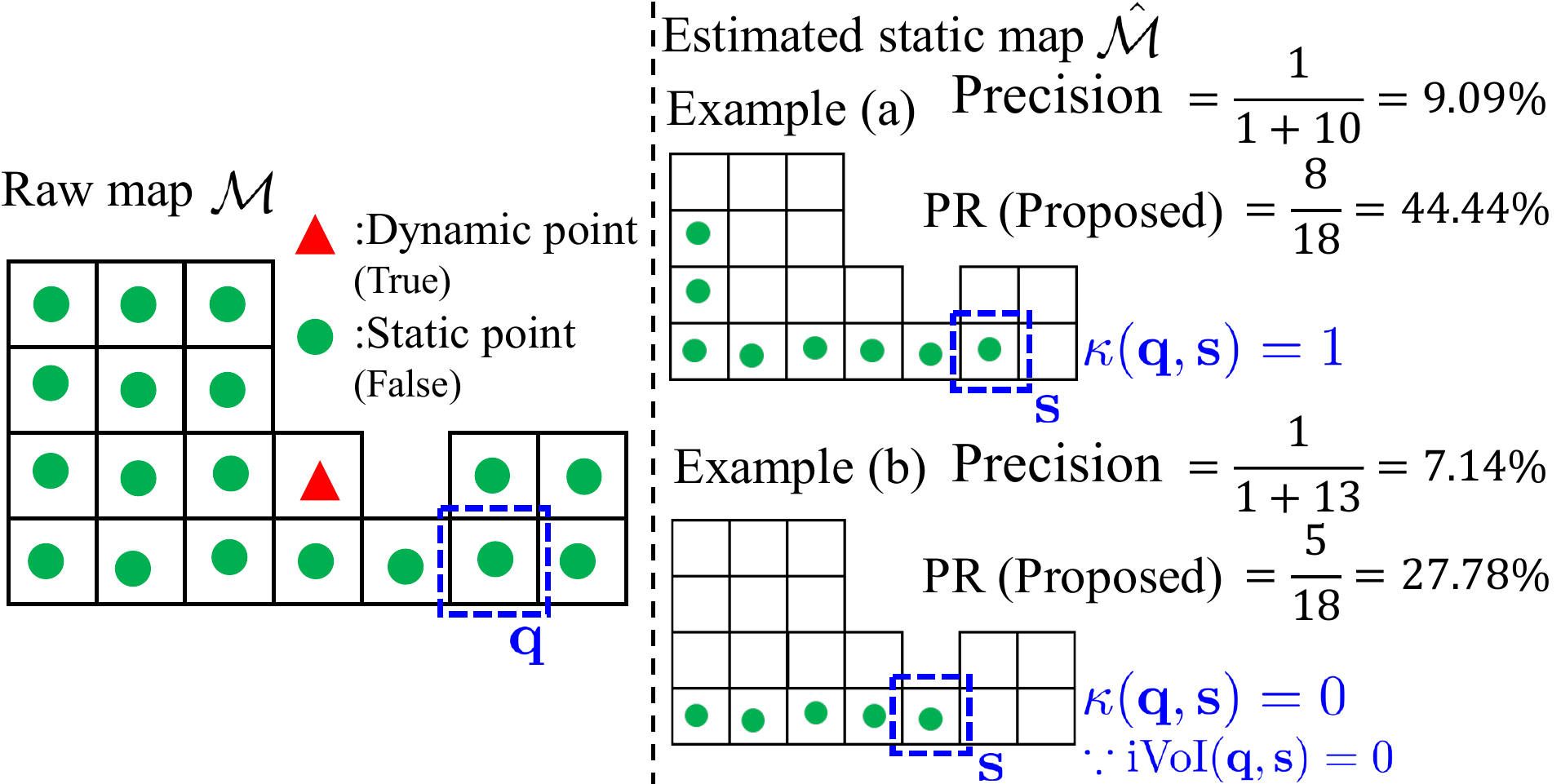}
    \end{subfigure}
    
    \caption{\textcolor{black}{Visual description of Preservation Rate. Preservation Rate is more appropriate to measure the performance of static-map building algorithms over Precision-Recall due to its higher sensitivity to the number of preserved static points.}}
    \label{fig:pp_pr }
\end{figure}

In addition, novel static map-oriented quantitative metrics called \textit{Rejection Rate (RR)} and \textit{Preservation Rate (PR)} are proposed. \textcolor{black}{An existing metric, namely, precision-recall \cite{kimremove}, is somewhat unsuitable for static map building tasks for the following reasons. First, as the number of static points to be erased (\textit{false positives, FP}) is much larger than the number of dynamic points (\textit{true positives, TP}), the precision is always too low. Thus it is not sensitive to relative changes as shown in Fig.~\ref{fig:pp_pr }. Second, use of different voxel sizes for different algorithms affects the number of FPs and TPs, i.e. precision-recall is a voxelization-variant metric.}


\textcolor{black}{The aforementioned metrics are defined as follows:}
\begin{itemize}
    \item PR: $\textcolor{black}{\frac{\text{\# of preserved static points}}{\text{\# of total static points on the raw map}}}$
    \item RR: $\textcolor{black}{1- \frac{\text{\# of preserved dynamic points}}{\text{\# of total dynamic points on the raw map}}}$.
\end{itemize}

\textcolor{black}{PR and RR are calculated voxel-wise. Here, as all different state-of-the-art methods use their own voxel size, we apply identical voxelization $\nu(\cdot)$ with a voxel size of 0.2 for static maps retrieved from the baseline models for fair comparison.}

\textcolor{black}{Let $\textbf{q} \in \nu(\mathcal{M})$ and $\textbf{s} \in \nu(\hat{\mathcal{M}})$ where $\textbf{s}$ is the nearest point to $\textbf{q}$. Then, the preservation function $\kappa(\cdot)$ is defined as follows:}
\begin{equation} 
\textcolor{black}{\kappa(\textbf{q}, \textbf{s})=\text{iVoI}(\textbf{q}, \textbf{s}) \land \psi(\textbf{q}, \textbf{s})}
\end{equation}
\textcolor{black}{where iVoI means inter Voxel Inclusion and is a boolean function that returns $\mathbf{true}$ if two points are in the same voxel, and $\mathbf{false}$ otherwise. Likewise, $\psi(\cdot)$ returns $\mathbf{true}$ if both points are static or dynamic and $\mathbf{false}$ otherwise. Fig.~\ref{fig:pp_pr } illustrates the examples of PR calculation. By measuring the joint frequency, our metrics reflect both the number of correct points and the inter-spatial relationships.}


\section{Results and Discussion} 
\subsection{Impact of the Ground Threshold} \label{sec:ground_thr}
\begin{figure}[h]
	\centering
	\begin{subfigure}[b]{0.24\textwidth}
		\includegraphics[width=1.0\textwidth]{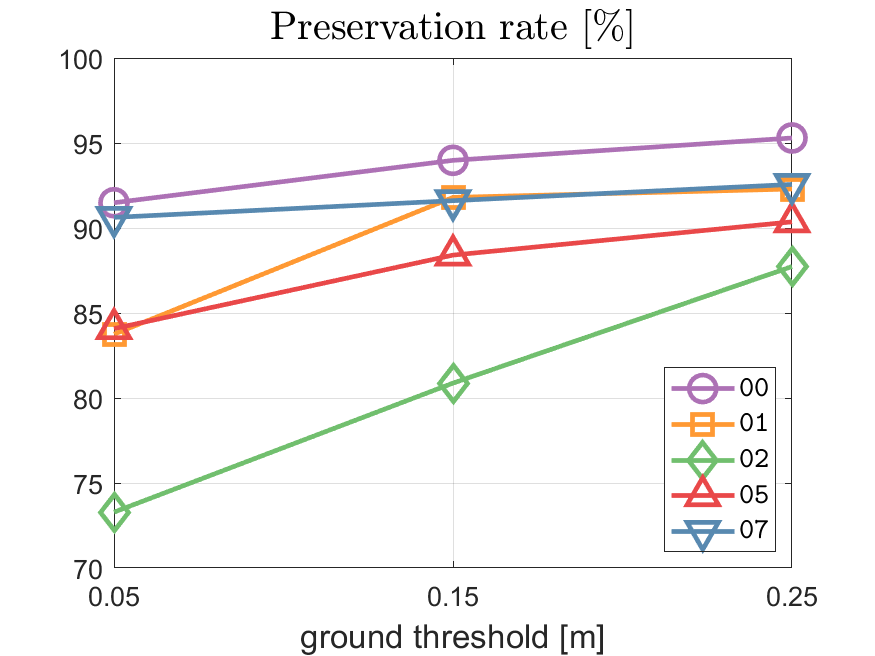}
		\label{fig:preservation ratio}
	\end{subfigure}
	\begin{subfigure}[b]{0.24\textwidth}
		\includegraphics[width=1.0\textwidth]{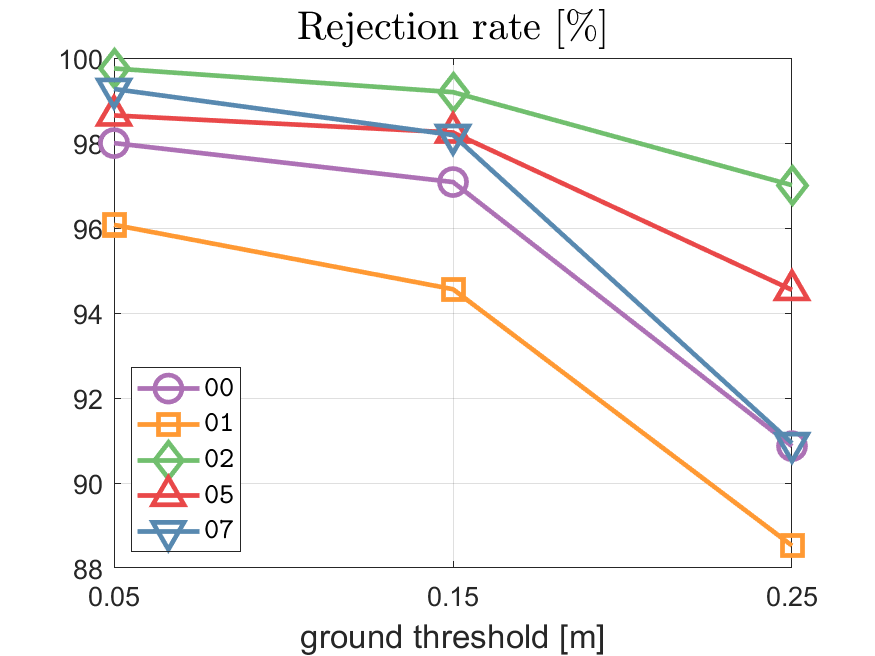}
		\label{fig:rejection ratio}
	\end{subfigure}
	\caption{\textcolor{black}{Performance changes based on \textcolor{black}{the} ground threshold~$\tau_{g}$. The higher the Preservation Rate and Rejection Rate, the better.}}
	\label{fig:ground_thr}
\end{figure}

First of all, the impact of the ground threshold $\tau_g$ on the proposed metrics was examined. As can be shown in Fig.~\ref{fig:ground_thr}, increasing $\tau_g$ led to improvement of the Preservation Rate because it \textcolor{black}{allowed more static points to be reverted}. Unfortunately, higher $\tau_g$ also allowed more dynamic points that are closer to the ground to be classified as ground. As a result, higher $\tau_g$ yielded a lower Rejection Rate. In most cases, Preservation Rates showed sharper increments from 0.05 to 0.15, \textcolor{black}{compared} to that from 0.15 to 0.25, whereas Rejection Rates showed sharper decrements from 0.15 to 0.25, compared to that from 0.05 to 0.15. Therefore, we 
\textcolor{black}{can conclude} that the ground threshold value of 0.15 yielded the most reasonable plane-fitting performance throughout the experiment.

\subsection{Comparison of R-GPF with Ground Plane Fitting} 

\begin{figure}[h]
    \centering
    \begin{subfigure}[b]{.95\linewidth}
        \includegraphics[width=\linewidth]{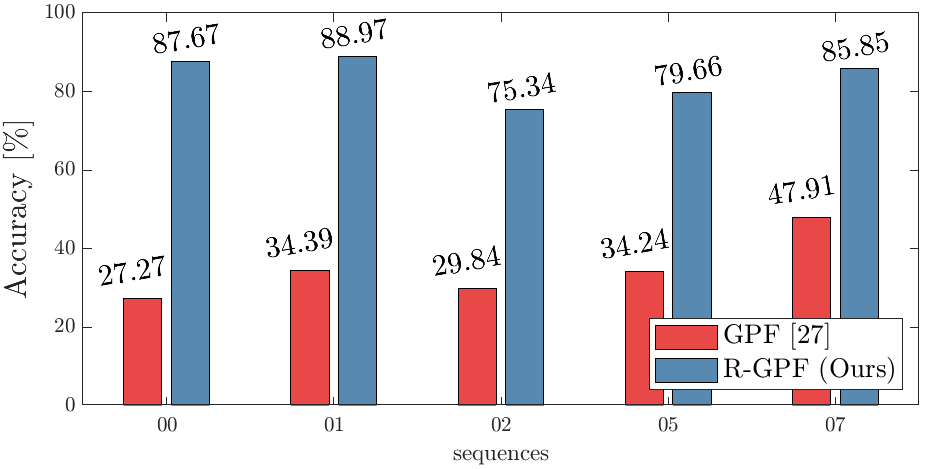}
    \end{subfigure}
    \caption{\textcolor{black}{Mean accuracy of static point retrieval.}} 
    \label{fig:gpf_vs_rgpf}
\end{figure}

\begin{table}[h]
    \centering
    \caption{Performance \textcolor{black}{comparison} with different ground plane fitting algorithms on SemanticKITTI dataset with $\tau_{g}=0.15$.}
    \begin{tabular}{clcccccc}
		\toprule
	Seq.   &   Input method & ${\mathbb{E}}(N_{s,g})$ & ${\mathbb{E}}(N_{d,g})$ & ${\mathbb{E}}(\hat{N}_{s,g})$ & ${\mathbb{E}}(\hat{N}_{d,g})$ \\ \midrule
	\multirow{2}{*}	{\texttt{00}}     
	& GPF \cite{zermas2017fast} &  \multirow{2}{*}{29.49k}   &\multirow{2}{*}{0.42k} &  7.74k & \textbf{2.15} \\
    & R-GPF (Ours)   & & & \textbf{25.85k} & 40.69 \\ \midrule
    \multirow{2}{*}	{\texttt{01}}     
	& GPF \cite{zermas2017fast} &  \multirow{2}{*}{12.25k}   &\multirow{2}{*}{0.45k} &  3.93k & \textbf{14.59} \\
    & R-GPF (Ours)   & & & \textbf{10.90k} & 44.18 \\ \midrule
    \multirow{2}{*}	{\texttt{02}}     
	& GPF \cite{zermas2017fast} &  \multirow{2}{*}{54.07k}   &\multirow{2}{*}{1.04k} &  15.43k & \textbf{20.22}\\
    & R-GPF (Ours)   & & & \textbf{40.71k} & 238.22 \\ \midrule
    \multirow{2}{*}	{\texttt{05}}     
	& GPF \cite{zermas2017fast} &  \multirow{2}{*}{32.55k}   &\multirow{2}{*}{1.65k} & 10.09k & \textbf{29.79} \\
    & R-GPF (Ours)   & & & \textbf{25.71k} & 109.16 \\ \midrule
    \multirow{2}{*}	{\texttt{07}}     
	& GPF \cite{zermas2017fast} &  \multirow{2}{*}{11.71k}   &\multirow{2}{*}{1.58k} &  4.79k & \textbf{4.95} \\
    & R-GPF (Ours)   & & & \textbf{9.89k} & 56.11 \\
    \bottomrule
	\end{tabular}
	\label{table:gpf_vs_rgpf}
\end{table}


\begin{figure*}[ht]
	\centering
	\begin{subfigure}[b]{0.195\textwidth}
		\includegraphics[width=1.0\textwidth]{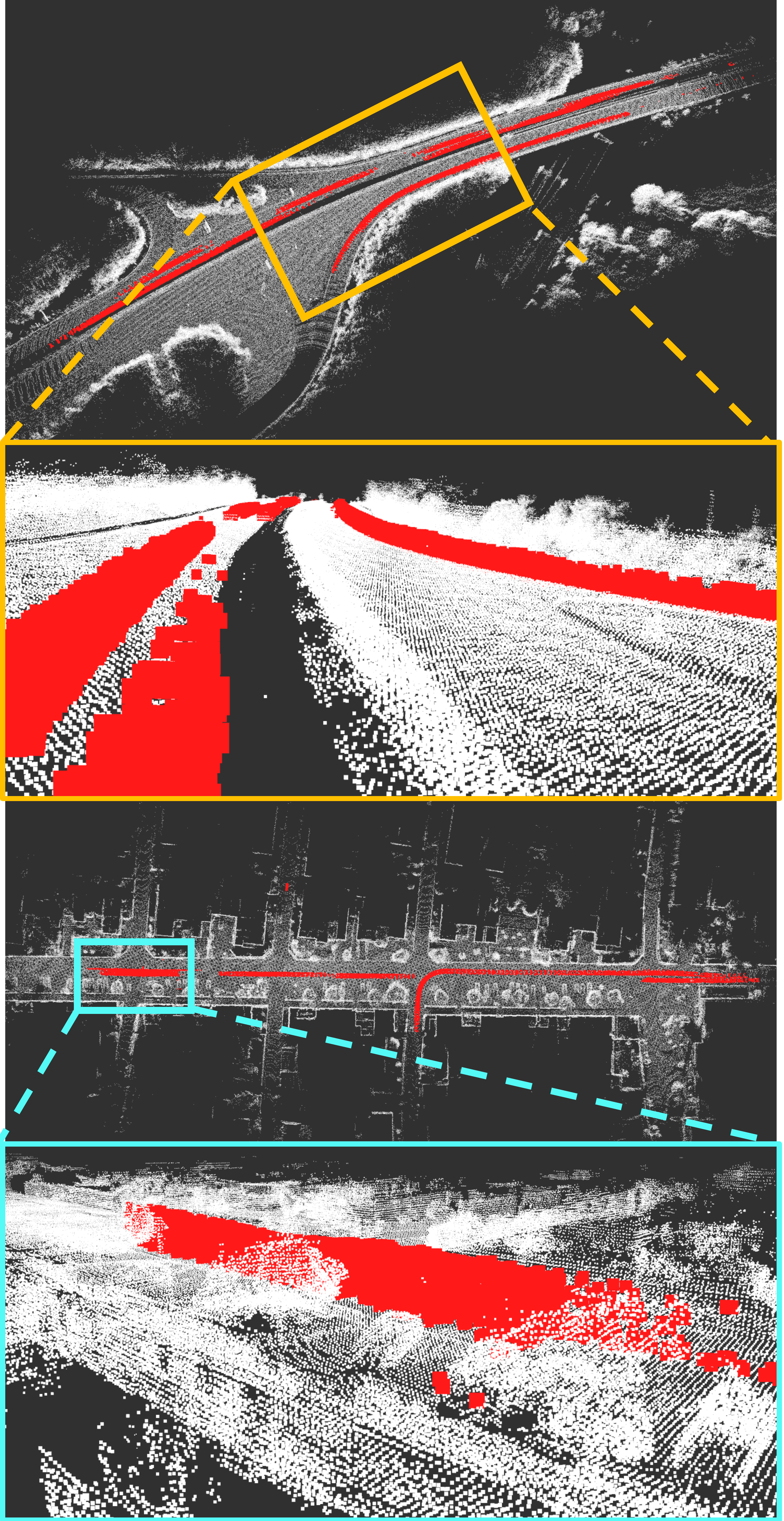}
		\caption{Original map}
		\label{fig:kaist_entrance}
	\end{subfigure}
	\begin{subfigure}[b]{0.195\textwidth}
		\includegraphics[width=1.0\textwidth]{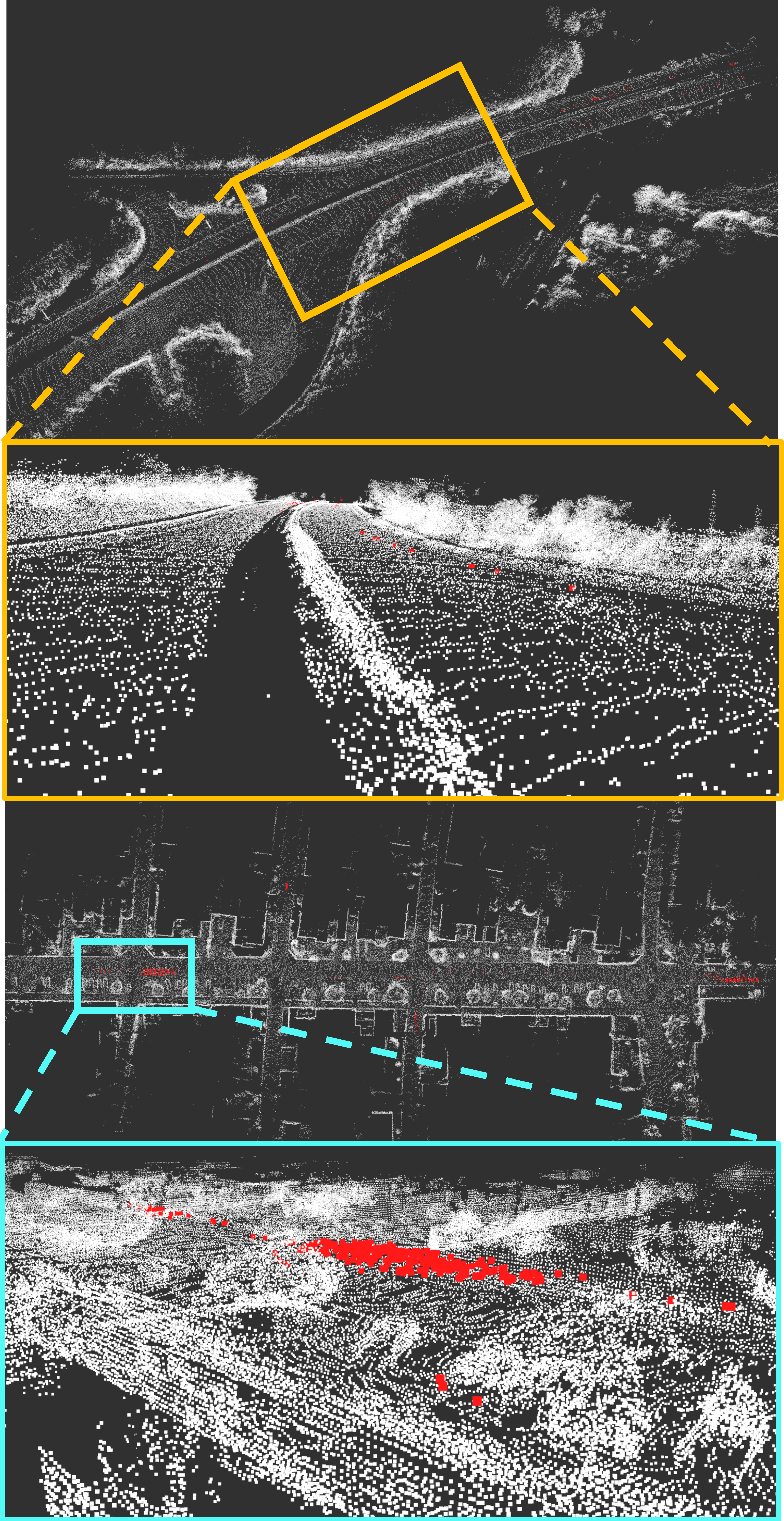}
		\caption{OctoMap \cite{wurm2010octomap}}
		
	\end{subfigure}
	\begin{subfigure}[b]{0.195\textwidth}
		\includegraphics[width=1.0\textwidth]{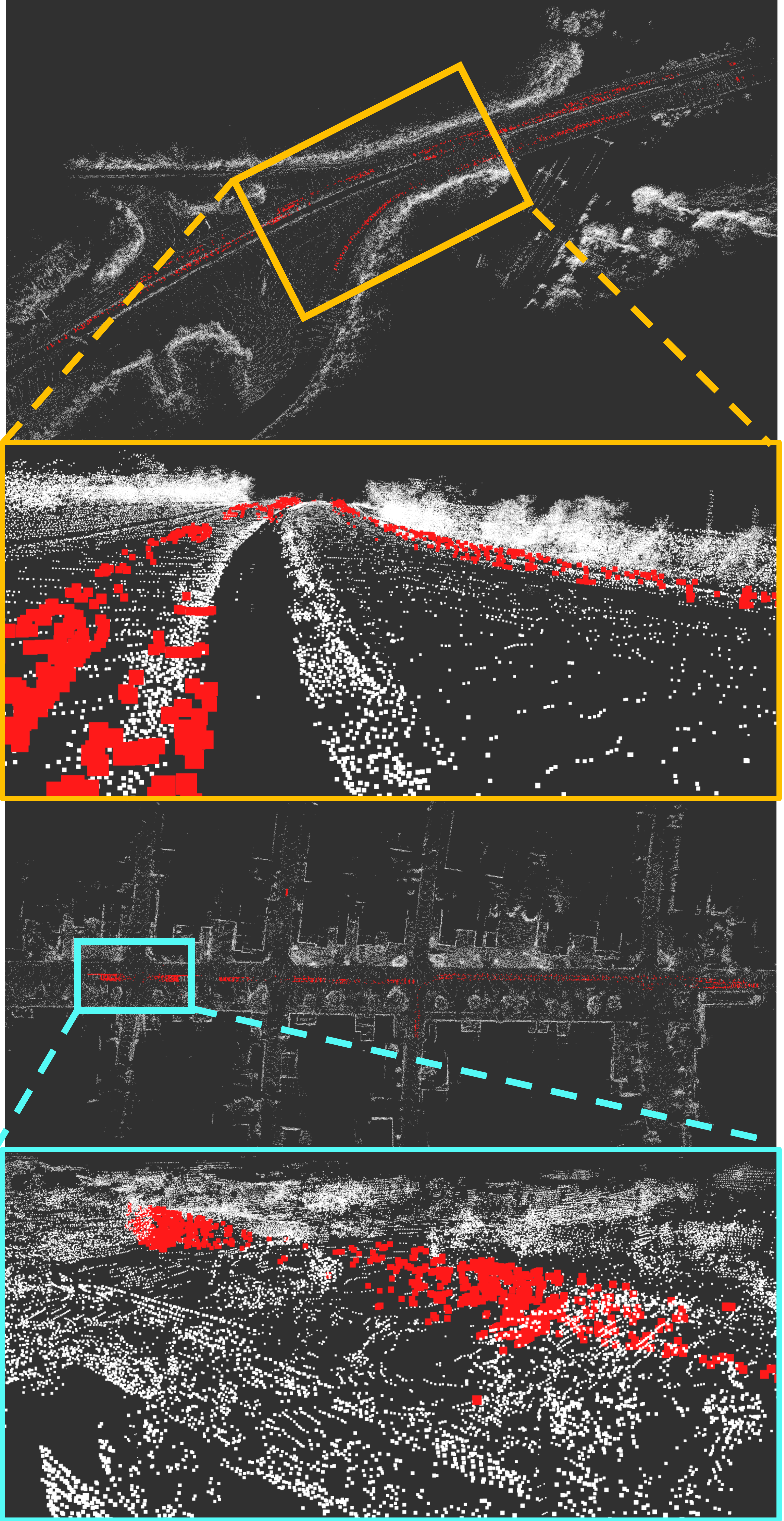}
		\caption{Peopleremover \cite{schauer2018peopleremover}}
		
	\end{subfigure}
	\begin{subfigure}[b]{0.195\textwidth}
		\includegraphics[width=1.0\textwidth]{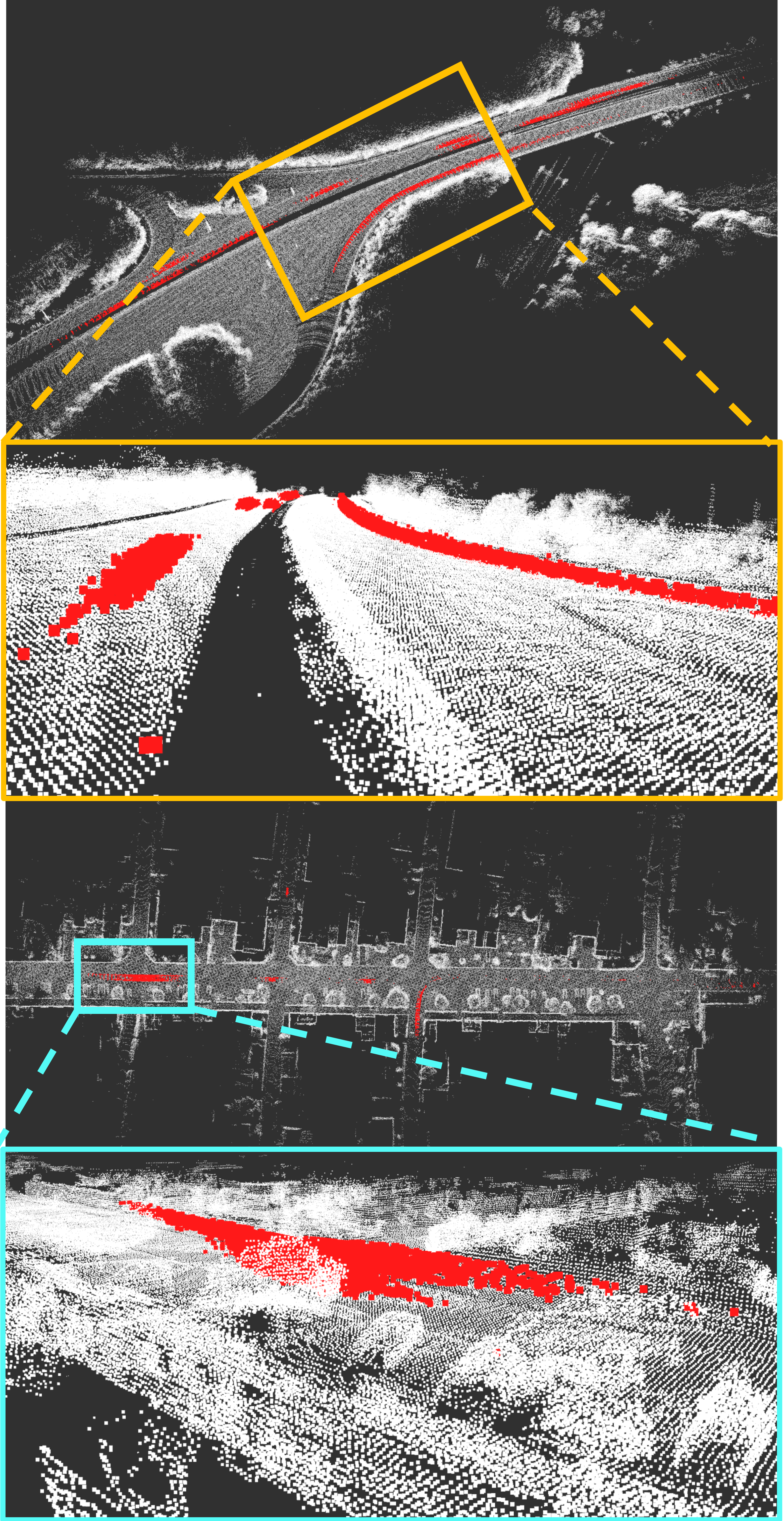}
		\caption{Removert \cite{kimremove}}
		
	\end{subfigure}
	\begin{subfigure}[b]{0.195\textwidth}
		\includegraphics[width=1.0\textwidth]{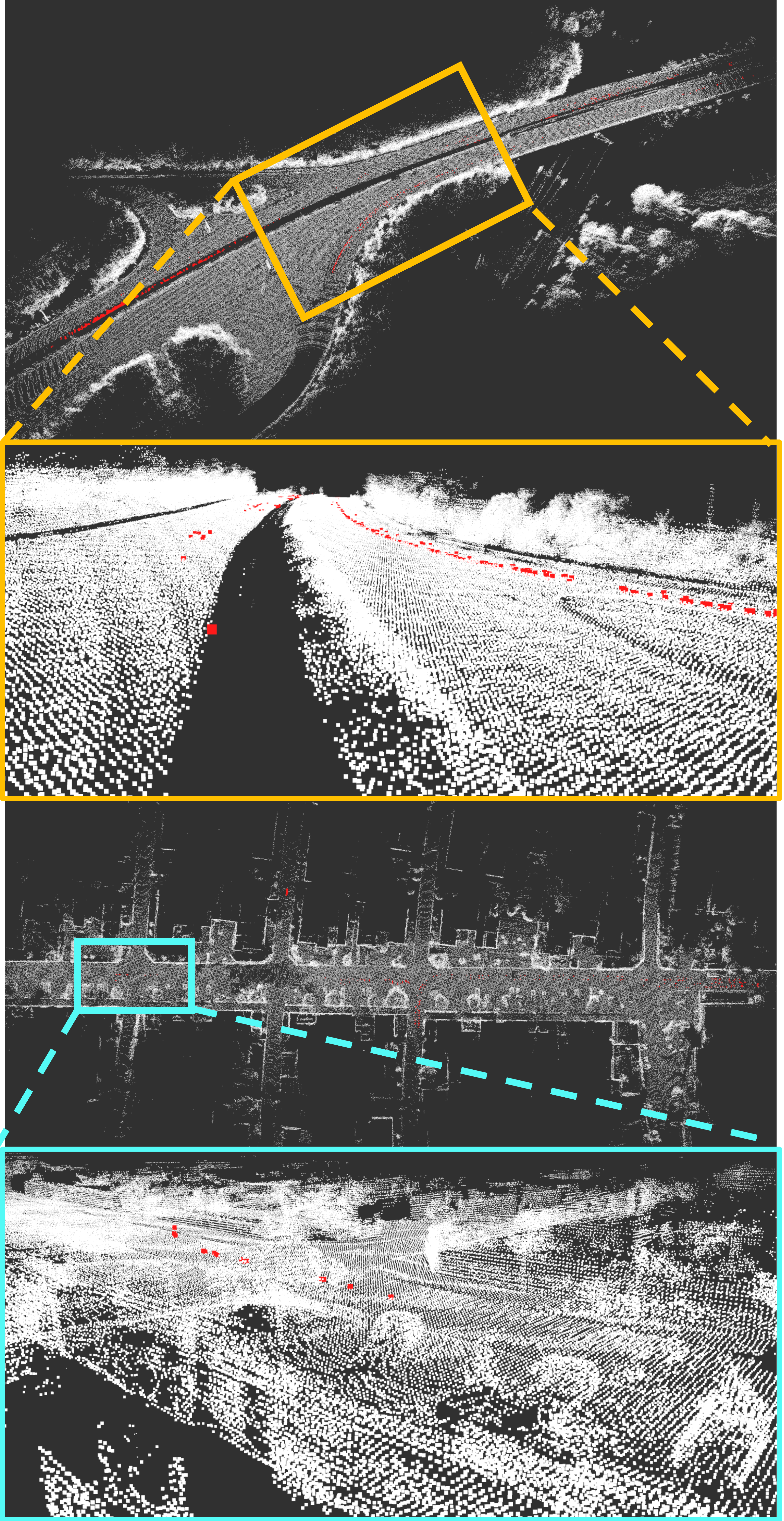}
		\caption{Proposed}
		
	\end{subfigure}
	\caption{\textcolor{black}{Comparison of static map generation results produced by the proposed method and state-of-the-arts methods on Semantic KITTI dataset (T-B): sequences \texttt{01} and \texttt{05}. Red cloud points indicate dynamic objects and the fewer red points, the better (best viewed in color).}}
	\label{fig:semantic_kitti_01_05}
\end{figure*}

\begin{figure*}[ht]
	\centering
	\begin{subfigure}[b]{0.195\textwidth}
		\includegraphics[width=1.0\textwidth]{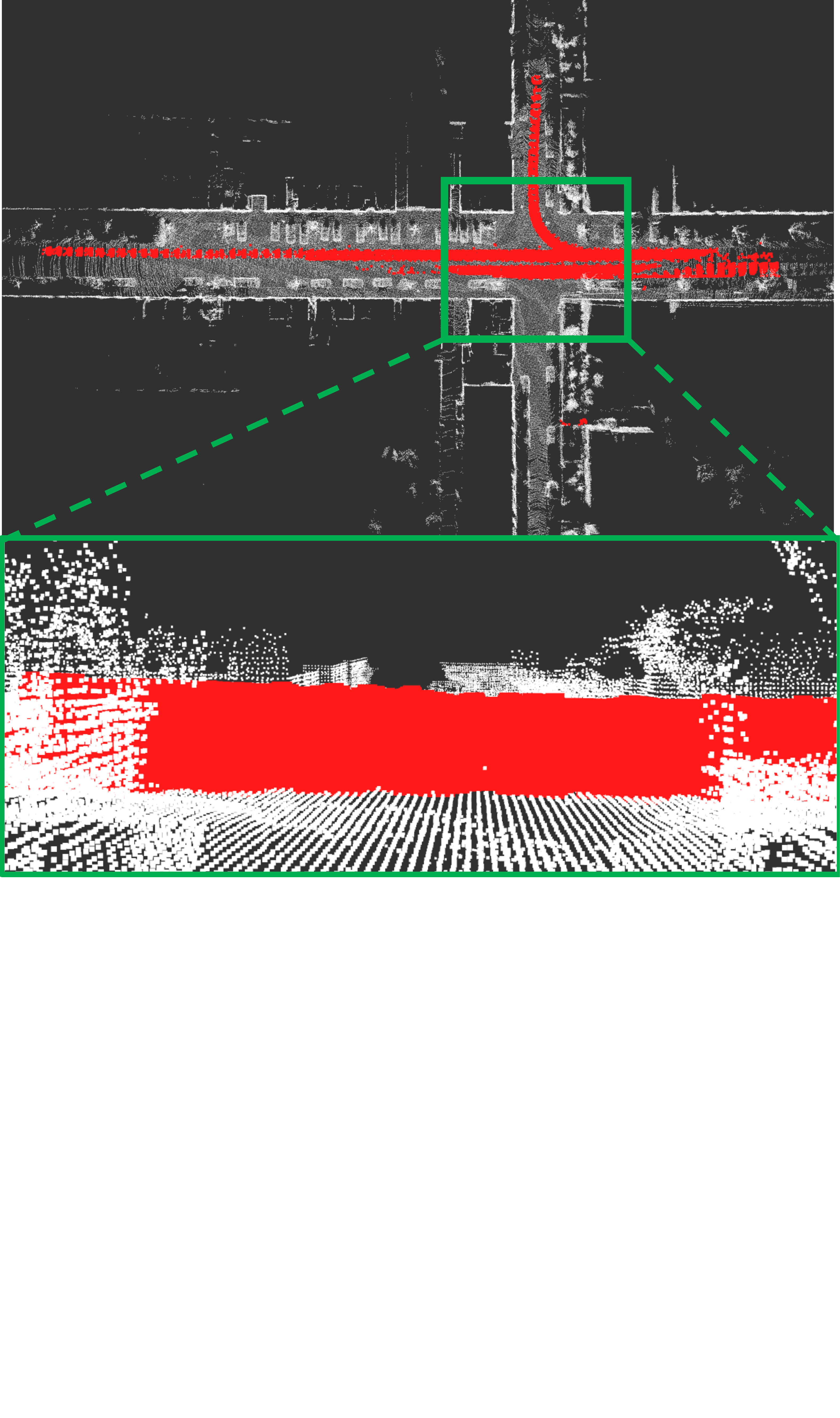}
		\caption{Original map}
		
	\end{subfigure}
	\begin{subfigure}[b]{0.195\textwidth}
		\includegraphics[width=1.0\textwidth]{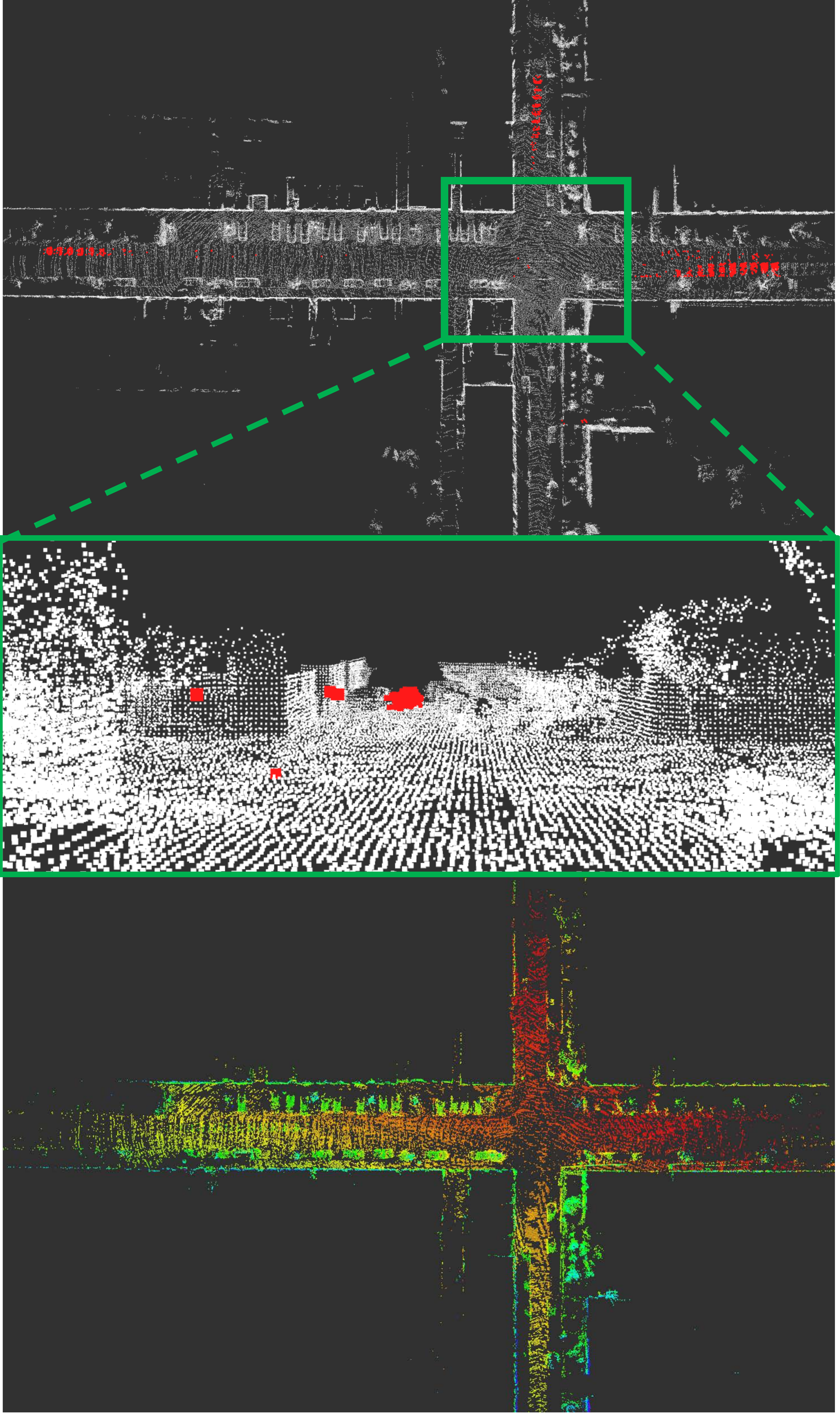}
		\caption{OctoMap \cite{wurm2010octomap}}
		
	\end{subfigure}
	\begin{subfigure}[b]{0.195\textwidth}
		\includegraphics[width=1.0\textwidth]{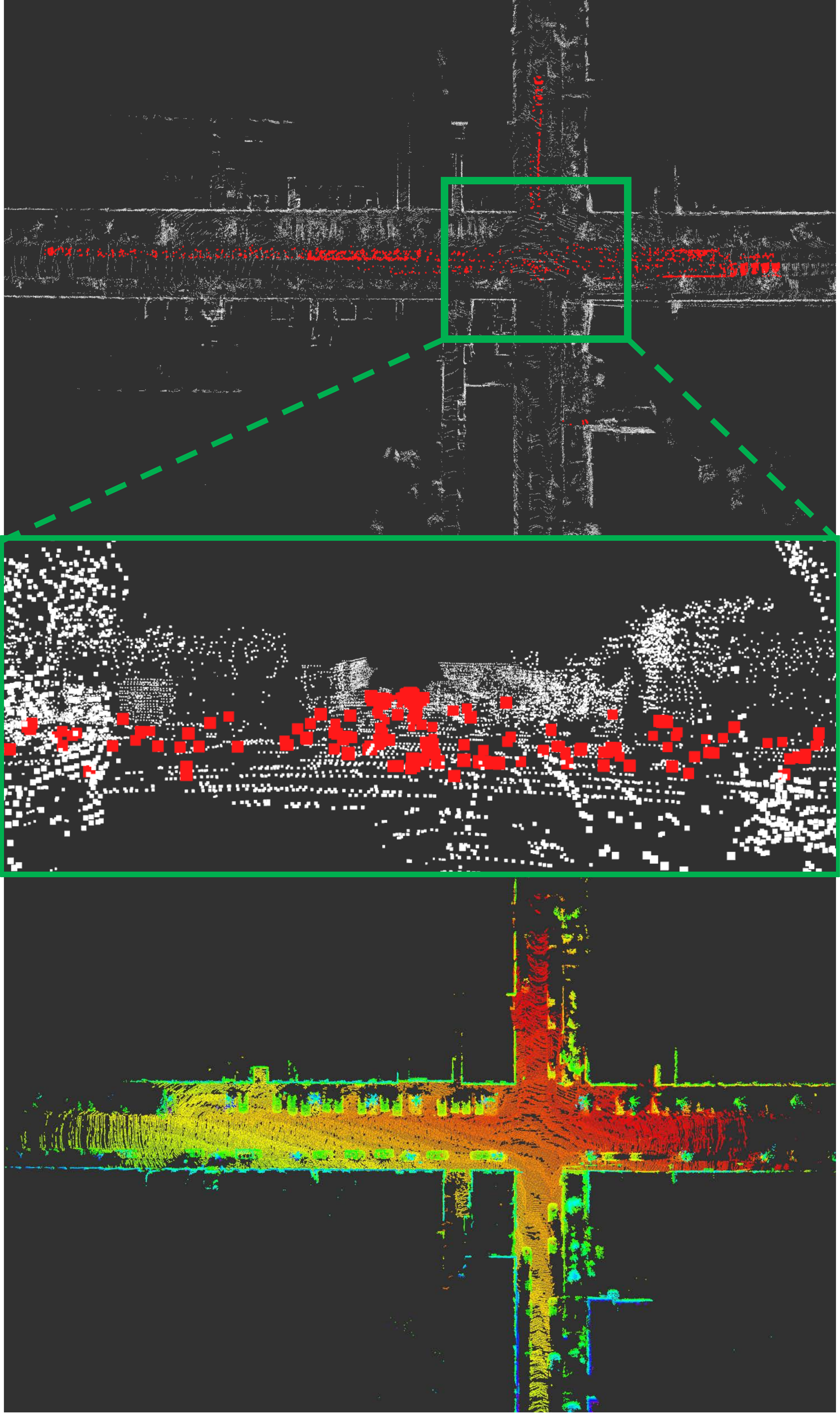}
		\caption{Peopleremover \cite{schauer2018peopleremover}}
		
	\end{subfigure}
	\begin{subfigure}[b]{0.195\textwidth}
		\includegraphics[width=1.0\textwidth]{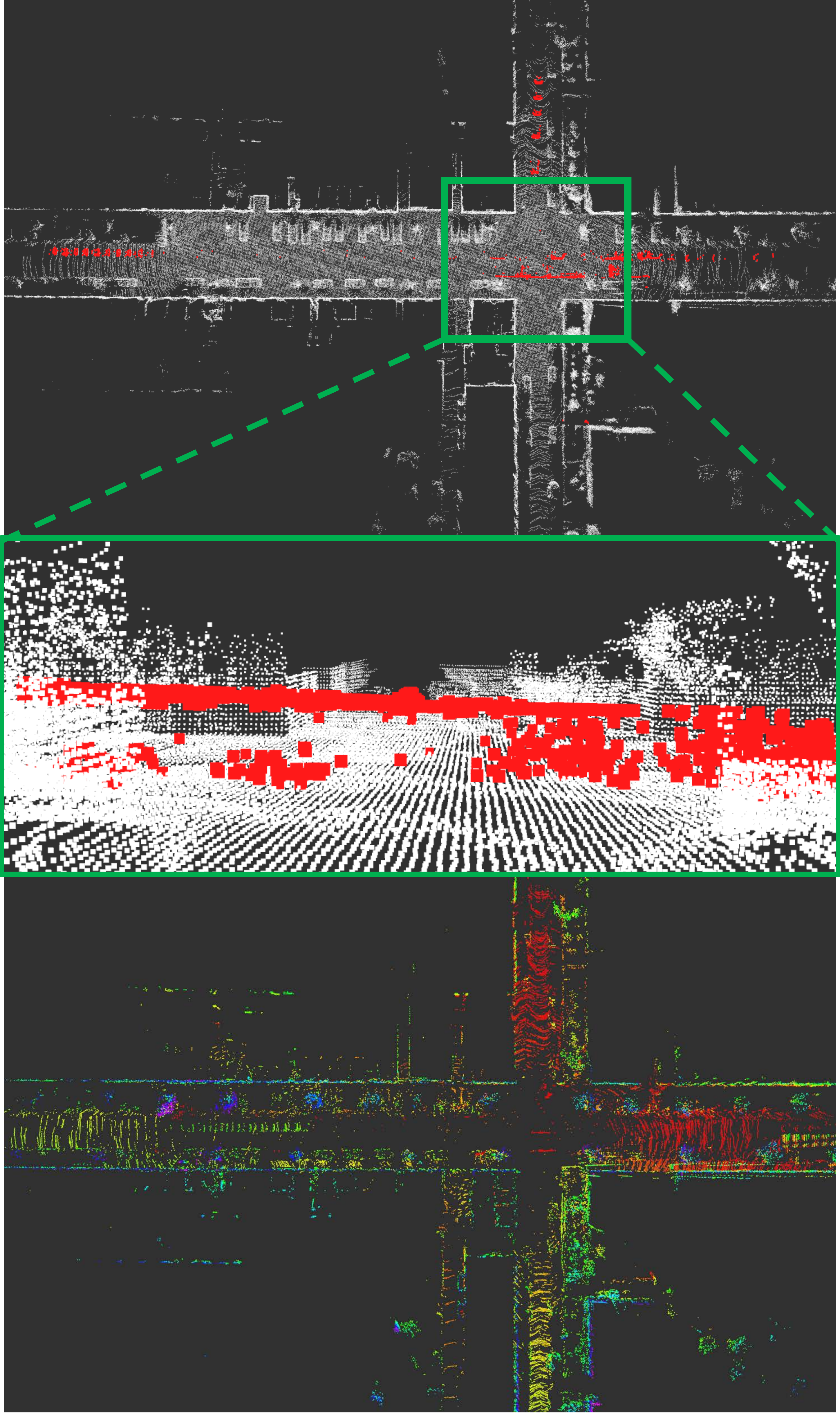}
		\caption{Removert \cite{kimremove}}
		
	\end{subfigure}
	\begin{subfigure}[b]{0.195\textwidth}
		\includegraphics[width=1.0\textwidth]{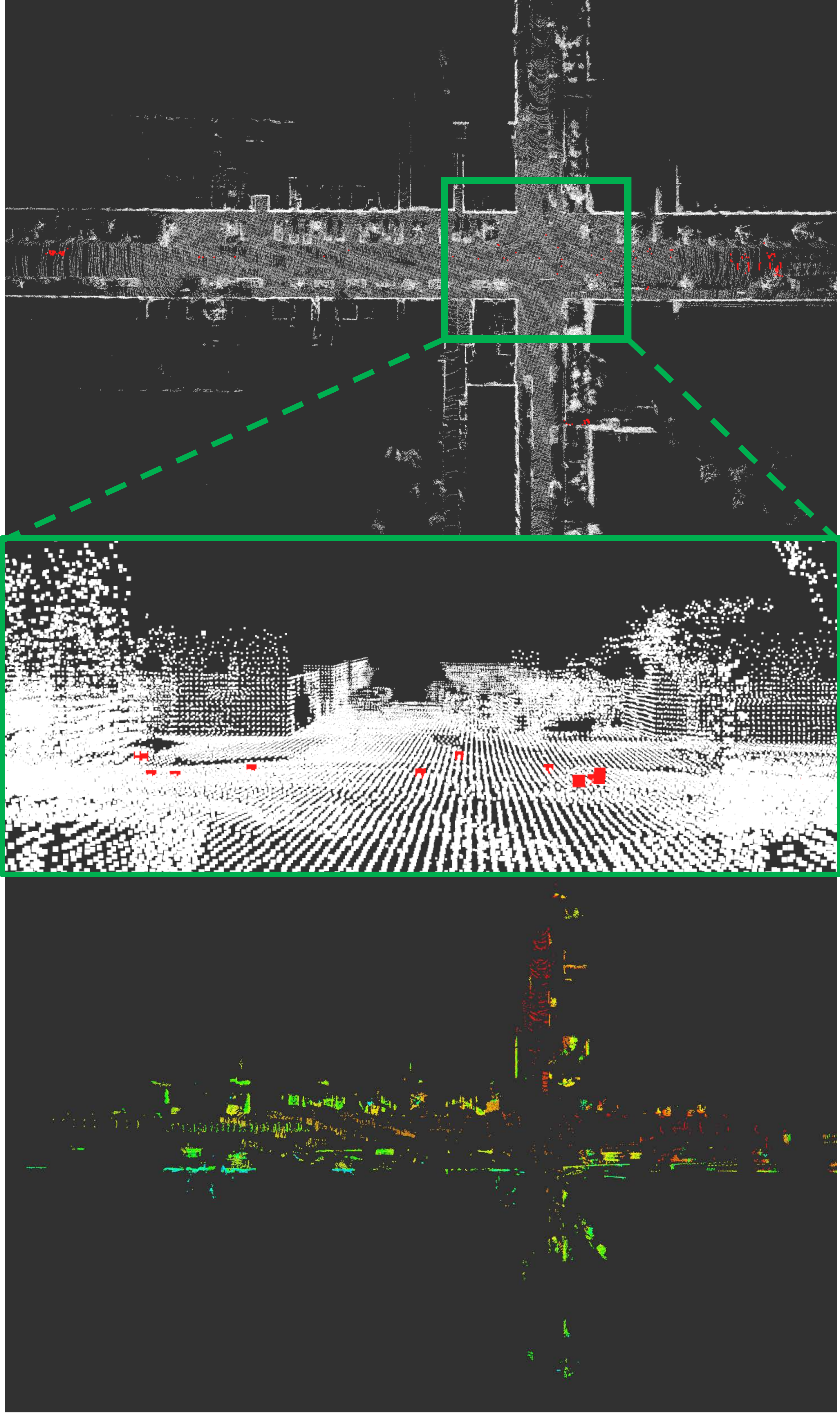}
		\caption{Proposed}
		
	\end{subfigure}
	\caption{\textcolor{black}{Comparison of static map generation results obtained with the proposed method and state-of-the-arts on sequence \texttt{07} of Semantic KITTI dataset (T-B): results of static map and lost static points (\textit{false positives}) during moving object removal. For both red cloud points and lost static points, the fewer the better (best viewed in color).}}
	\label{fig:semantic_kitti_07}
\end{figure*}


As shown in Table~\ref{table:gpf_vs_rgpf} and Fig.~\ref{fig:gpf_vs_rgpf} \textcolor{black}{where} $\mathbb{E}(\cdot)$ denotes expectation; $N_{s,g}$ and $N_{d,g}$ denote the total numbers of static and dynamic points of all potentially dynamic bins; and $\hat{N}_{s,g}$ and $\hat{N}_{d,g}$ denote the total number of static and dynamic points selected by an algorithm, respectively, our R-GPF yielded higher performance in static point retrieval over GPF \cite{zermas2017fast} in various urban environments. In particular, the difference in performance is the largest for the sequence \texttt{00}, where many curbs are placed to form non-planar floors, from which we can infer that R-GPF leverages the advantage of bin-wise operation over the non-planar floor as explained in \textcolor{black}{Section~\rom{2}.\textit{E}}. \textcolor{black}{Note that R-GPF reverts a few more dynamic points, i.e. higher ${\mathbb{E}}(\hat{N}_{d,g})$, because R-GPF is likely to consider the underside of dynamic points, such as the contact point of a vehicle wheel, as inliers.} However, R-GPF was originally intended to return more static points to the map risking a few dynamic points, which is motivated by the Revert stage of Removert \cite{kimremove}. Besides, these dynamic points are so close to the ground that they have an insignificant impact on localization or navigation performance. Therefore, \textcolor{black}{the effect of} these undesirable points could be negligible.

\subsection{Comparison with the State-of-the-Art Methods} 

\begin{table}[h]
	\centering
	\caption{\textcolor{black}{Comparison with state-of-the-art methods on the SemanticKITTI dataset. PR: Preservation Rate, RR: Rejection Rate (see Section \rom{3}.\textit{B}).}}
	\begin{tabular}{llccc}
	\toprule
	Seq. & Method  & \multicolumn{1}{c}{\begin{tabular}[c]{@{}c@{}} PR [\%] \end{tabular}}  & \multicolumn{1}{c}{\begin{tabular}[c]{@{}c@{}} RR [\%] \end{tabular}} & $\text{F}_{1}$ score  \\ \midrule
		\multirow{6}{*}{\texttt{00}}&\multirow{1}{*}{OctoMap - \texttt{0.05} \cite{wurm2010octomap}} &  76.731  & 99.124 &  0.865  \\ 
		&\multirow{1}{*}{OctoMap - \texttt{0.2}  \cite{wurm2010octomap}} & 34.568  & \textbf{99.979} &  0.514  \\ 
		
		&\multirow{1}{*}{Peopleremover  \cite{schauer2018peopleremover}} & 37.523  & 89.116 &  0.528  \\ 
		
		&\multirow{1}{*}{Removert - \texttt{RM3} \cite{kimremove}} & 85.502  & 99.354 & 0.919  \\ 
		&\multirow{1}{*}{Removert - \texttt{RM3+RV1} \cite{kimremove}} & 86.829  & 90.617 &  0.887  \\ 
		& ERASOR (Ours) & \textbf{93.980}  & 97.081 &  \textbf{0.955} \\ \midrule
		\multirow{6}{*}{\texttt{01}}&\multirow{1}{*}{OctoMap - \texttt{0.05} \cite{wurm2010octomap}} & 53.163  & 99.663 &  0.693  \\ 
		&\multirow{1}{*}{OctoMap - \texttt{0.2}  \cite{wurm2010octomap}} & 20.777  & \textbf{99.863} &  0.344  \\ 
		
		&\multirow{1}{*}{Peopleremover  \cite{schauer2018peopleremover}} & 36.349  & 93.116 &  0.523  \\ 
		
		&\multirow{1}{*}{Removert - \texttt{RM3} \cite{kimremove}} &  94.221  & 93.608 & \textbf{0.939}  \\ 
		&\multirow{1}{*}{Removert - \texttt{RM3+RV1} \cite{kimremove}} &  \textbf{95.815}  & 57.077 &  0.715  \\ 
		& ERASOR (Ours) & 91.487 & 95.383  & 0.934  \\ \midrule
		\multirow{6}{*}{\texttt{02}}&\multirow{1}{*}{OctoMap - \texttt{0.05} \cite{wurm2010octomap}} & 54.112  & 98.769 &  0.699  \\ 
		&\multirow{1}{*}{OctoMap - \texttt{0.2}  \cite{wurm2010octomap}} & 23.746  & \textbf{99.792} &  0.384  \\ 
		
		&\multirow{1}{*}{Peopleremover  \cite{schauer2018peopleremover}} & 29.037  & 94.527 &  0.444  \\ 
		
		&\multirow{1}{*}{Removert - \texttt{RM3} \cite{kimremove}}  & 76.319  & 96.799 &  0.853  \\ 
		&\multirow{1}{*}{Removert - \texttt{RM3+RV1} \cite{kimremove}} & 83.293  & 88.371 &  0.858  \\ 
		
		& ERASOR (Ours)   & \textbf{87.731}  & 97.008  & \textbf{0.921} \\ \midrule
		\multirow{6}{*}{\texttt{05}}&\multirow{1}{*}{OctoMap - \texttt{0.05} \cite{wurm2010octomap}}  & 76.341  & 96.785 &  0.854  \\ 
		&\multirow{1}{*}{OctoMap - \texttt{0.2}  \cite{wurm2010octomap}} & 33.904  & \textbf{99.882} &  0.506  \\ 
		
		&\multirow{1}{*}{Peopleremover  \cite{schauer2018peopleremover}} & 38.495  & 90.631 &  0.540  \\ 
		
		&\multirow{1}{*}{Removert - \texttt{RM3} \cite{kimremove}} & 86.900  & 87.880 &  0.874  \\ 
		&\multirow{1}{*}{Removert - \texttt{RM3+RV1} \cite{kimremove}} & 88.170  & 79.981 & 0.839  \\ 
		
		& ERASOR (Ours) & \textbf{88.730}  & 98.262  & \textbf{0.933}  \\ \midrule
		\multirow{6}{*}{\texttt{07}}&\multirow{1}{*}{OctoMap - \texttt{0.05} \cite{wurm2010octomap}} & 77.838  & 96.938 &  0.863  \\ 
		&\multirow{1}{*}{OctoMap - \texttt{0.2}  \cite{wurm2010octomap}} & 38.183 & \textbf{99.565}  &  0.552  \\ 
		
		&\multirow{1}{*}{Peopleremover  \cite{schauer2018peopleremover}} & 34.772  & 91.983 &  0.505  \\ 
		
		&\multirow{1}{*}{Removert - \texttt{RM3} \cite{kimremove}} & 80.689  & 98.822 &  0.888 \\ 
		&\multirow{1}{*}{Removert - \texttt{RM3+RV1} \cite{kimremove}} & 82.038  & 95.504 &  0.883  \\ 
		
		& ERASOR (Ours)  & \textbf{90.624}  & 99.271  & \textbf{0.948}  \\ \bottomrule
	\end{tabular}
	\label{table:kitti_comparison}
\end{table}

ERASOR was quantitatively compared with state-of-the-art methods, namely, \textcolor{black}{OctoMap\footnote{https://octomap.github.io/} \cite{wurm2010octomap}, Peopleremover\footnote{http://threedtk.de} \cite{schauer2018peopleremover}, and Removert\footnote{https://github.com/irapkaist/removert} \cite{kimremove}. We leveraged the open source implementations for the experiment. In Table~\ref{table:kitti_comparison}, \texttt{0.05} and \texttt{0.2} denote the processing voxel sizes of OctoMap.} \texttt{RM3} denotes the results after three Removal stages with per-pixel resolutions \cite{kimremove} of 0.4$^\circ$, 0.45$^\circ$, and 0.5$^\circ$. \texttt{RM3-RV1} means the result of \texttt{RM3} followed by a Revert stage with the resolution per pixel of 0.55$^\circ$.

The state-of-the-art methods give an exquisite static map, filtering out the most dynamic points. However, our ERASOR exhibits noticeable improvements over the aforementioned limitations of both ray tracing-based and visibility approaches, as illustrated in Fig.~\ref{fig:visibility_limitation}. Note that our methodology is fundamentally free from visibility.

\textcolor{black}{In particular, the static map generation results by the state-of-the-arts on the sequence \texttt{05} and \texttt{07} reported that there are still some remaining dynamic points on the top part of the bus due to the limitation explained in Fig.~\ref{fig:visibility_limitation}(c), as shown in the last row of Fig.~\ref{fig:semantic_kitti_01_05} and Fig.~\ref{fig:semantic_kitti_07}.} That is, the object was too large and too close to the query coordinate; thus, the remaining dynamic points were in an invalid range to be checked for ray tracing or visibility. In contrast, our proposed method successfully removes most of the dynamic traces. Because our approach checks the scan ratio, even the points existing in the invalid range of visibility can be removed.

\textcolor{black}{Although OctoMap rejects most of the dynamic points as shown in Fig.~\ref{fig:semantic_kitti_01_05} and Fig.~\ref{fig:semantic_kitti_07}, it falsely removes large parts of static points, which results in the lowest PR as shown in Table~\ref{table:kitti_comparison}. The last row of Fig.~\ref{fig:semantic_kitti_07} visualizes the lost static points, i.e. \textit{false positives}, which implies that the OctoMap and Peopleremover suffer from motion ambiguity and unintentionally loses a relatively larger number of static points compared to our proposed method.}

Therefore, our proposed method shows promising PR and $\text{F}_1$ scores, as shown in Table~\ref{table:kitti_comparison}, in all environments except for \texttt{01}. The cause of lower PR in this frame is presumed to be the removal of vegetation far from the highway because once vegetation is partially observed, and yields a small scan ratio.

\subsection{Algorithm speed} 
\textcolor{black}{In addition, our proposed method shows the fastest performance for one iteration, as reported in Table~\ref{table:kitti_speed}. Although the time complexity of Removert \cite{kimremove} is $O(n)$ and that of our ERASOR is $O(n\log{n})$, Removert requires multiple projections on the spherical image plane and assigns staticity to each point, while ERASOR is a bin-wise approach; thus, it removes the dynamic points with one-shot via the scan ratio. Therefore, our method is at least ten times faster than other methods.}

\begin{table}[h]
	\centering
	\caption{\textcolor{black}{Runtime per iteration on sequence \texttt{01} of SemanticKITTI dataset.}}
	\begin{tabular}{lc}
	\toprule
	Method & Runtime/iteration [s]   \\ \midrule
	OctoMap \cite{wurm2010octomap} & 1.077 \\
	Peopleremover \cite{schauer2018peopleremover} & 1,000 \\
	Removert \cite{kimremove} & 0.8307  \\
	ERASOR (Ours) & \textbf{0.0732}  \\  \bottomrule
	\end{tabular}
	\label{table:kitti_speed}
\end{table}

\section{Conclusion}
In this study, a novel static map \textcolor{black}{building} method, ERASOR, has been proposed. Our method was tested on the SemanticKITTI dataset both quantitatively and qualitatively. As a result, our method was proved to overcome the limitations of ray tracing-based methods and visibility-based methods, and it shows promising results in various environments. Our method provides a static map with few dynamic objects, \textcolor{black}{which helps improve the navigation and localization tasks of mobile robots.} In future works, we plan to improve pseudo occupancy or to devise a spatial descriptor that is more robust to other kinds of noise and is suitable for more sophisticated dynamic object removal.

\section*{Acknowledgment}

Above all things, we thank G. Kim for making his code available and sharing the results of Removert. Especially, we thank B. Lee for establishing error metrics. 

\ifCLASSOPTIONcaptionsoff
  \newpage
\fi




%

\bibliographystyle{IEEEtran}
\bibliography{./ral_erasor,./IEEEabrv}

\end{document}